%% file: arxiv.tex
\documentclass{article}

 \usepackage[preprint]{neurips_2026}

\input{math_commands.tex}

\usepackage[utf8]{inputenc} 
\usepackage[T1]{fontenc}    
\usepackage{url}            
\usepackage{booktabs}       
\usepackage{graphicx}       
\usepackage{amsfonts}       
\usepackage{nicefrac}       
\usepackage{microtype}      
\usepackage{xcolor}         
\usepackage{colortbl}       
\usepackage{xspace}
\usepackage{tabularx}
\usepackage{array}
\usepackage{ragged2e}
\usepackage{amsmath}
\usepackage{algorithm}
\usepackage{algorithmic}
\usepackage{soul}
\usepackage{makecell}
\usepackage{multirow}
\usepackage{hyperref}       

\newcolumntype{L}[1]{>{\RaggedRight\arraybackslash}p{#1}}
\newcolumntype{Y}{>{\RaggedRight\arraybackslash}X}

\newcommand{\oureval}{\textup{\textsc{AgentCL}}\xspace}
\newcommand{\ourmethod}{\textup{\textsc{MemProbe}}\xspace}
\usepackage{etoc}
\usepackage{titletoc}
\newcommand\AppendixTOC{ \startcontents \printcontents{}{1}{\noindent\textbf{\Large Appendix Contents}\vspace{4pt}} }
\newcommand{\ann}[1]{\,{\footnotesize\textcolor{gray}{(#1)}}}

\title{\oureval: Toward Rigorous Evaluation of \\ Continual Learning in Language Agents}

\author{%
  \makebox[\textwidth][c]{%
  \begin{tabular}{@{}l@{}}
  \textbf{Yiheng Shu$^{1}$, 
  Bernal Jim\'enez Guti\'errez$^{2}$, 
  Saisri Padmaja Jonnalagedda$^{3}$, Yuguang Yao$^{3}$,} \\
  \textbf{
  Huan Sun$^{1}$,
  Yu Su$^{1}$} \\
  \normalfont
  $^{1}$The Ohio State University \quad
  $^{2}$Johns Hopkins University \quad
  $^{3}$Intuit AI Research \\
  \texttt{\{shu.251, sun.397, su.809\}@osu.edu}
  \end{tabular}%
  }%
}

\begin{document}

\maketitle
\setcounter{footnote}{0}

\begin{abstract}
Language agents spend substantial inference time solving individual tasks, yet the experience acquired in one episode is often underutilized in future episodes.
Continual learning expects an agent to accumulate experience across a stream of tasks, improve over time, and avoid interference from irrelevant experiences.
Unfortunately, existing benchmarks struggle to evaluate this problem rigorously.
Most efforts focus on retrieval and reasoning over long-context conversations or documents, while recent lifelong-adaptation benchmarks often stream existing datasets whose cross-task dependencies are unknown, making it difficult to understand what an agent learns and reuses over time.
This paper presents an evaluation framework \oureval for continual learning in agents, centered on distinguishable task streams and metrics for CL properties.
\oureval distinguishes three cross-task relations by the availability of reusable knowledge. 
In dependent streams, later tasks can reuse knowledge from earlier ones. In conventional streams, whether such reuse exists is unknown. 
Besides, independent tasks are held out with little transferable knowledge.
We use the benchmark to evaluate non-parametric memory designs for continual learning.
To diagnose how memory design choices affect continual learning, we develop \ourmethod, a probing method that stores interactions, insights, and skills, while filtering unreliable experiences during consolidation.
Empirical analysis across coding, deep research, and language understanding/reasoning tasks shows that conventional streams offer limited ability to distinguish memory designs, whereas dependent streams more clearly distinguish their plasticity.
Meanwhile, conventional streams and independent tasks often yield limited gains and can expose memory-induced degradation.
These results highlight the need for stronger memory designs that balance plasticity and stable reuse.\footnote{Data is available at \url{https://huggingface.co/datasets/osunlp/AgentCL}.}
\end{abstract}

\section{Introduction}

\begin{figure}[t]
  \centering
  \includegraphics[width=0.95\textwidth]{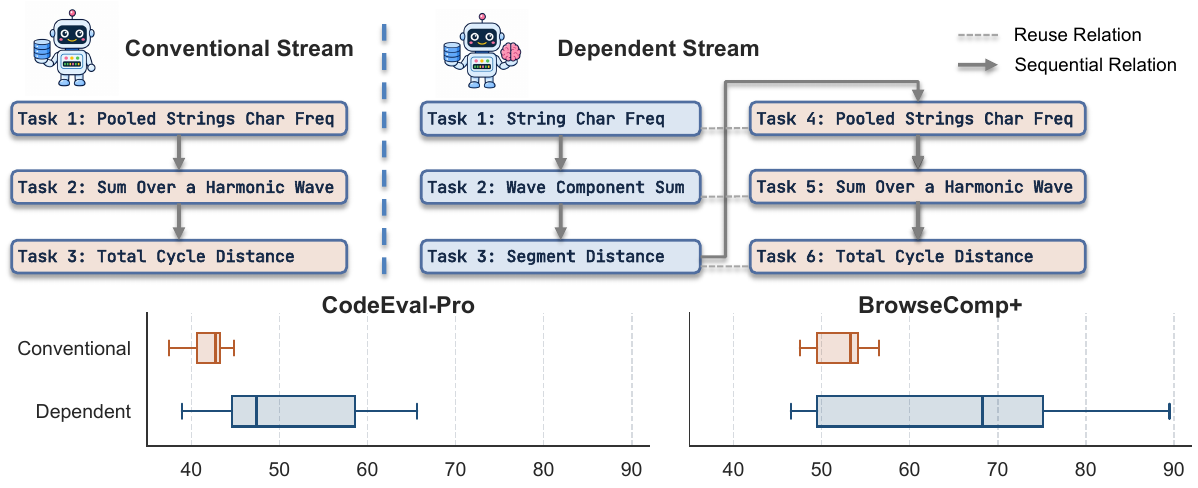}
  \caption{
  \oureval with task streams.
  \textit{Reuse relation} denotes a task pair in which the latter one can reuse the solution to the former one.
  \textit{Sequential relation} denotes the order of task execution.
\textbf{Top}: conventional streams order tasks from the same domain with unknown reuse relations, while dependent streams include reusable subtasks before related complex tasks.
\textbf{Bottom}: boxplots of average accuracy across evaluated methods on CodeEval-Pro and BrowseComp+, showing that dependent streams yield larger method separation than conventional streams for memory reuse.
  }
  \label{fig:eval}
  \vspace{-10pt}
\end{figure}

Language agents \citep{agent_tutorial} powered by large language models (LLMs) spend substantial inference time interacting with complex environments, yet the experience acquired in each episode is often underutilized before the next one begins. 
The fact that intensive inference does not automatically compound into long-term competence motivates the \textit{continual learning (CL) problem for language agents}: given a stream of tasks, can an agent accumulate reusable experience, successfully transfer it to future environments, and remain robust as its memory scales?

However, evaluating this capability challenges the conventional evaluation paradigms of machine learning.
Traditionally, LLMs are trained and evaluated in static batches under the i.i.d. assumption.
In contrast, agents often operate in a sequential setting with limited task experience, where cross-task relations are essential to measure continual learning.
Also, a delicate balance between \textit{plasticity} (adapting to new tasks) and \textit{stability} (retaining past knowledge) is required during the learning process \citep{cl_survey}.
Given these unique characteristics, we argue that a rigorous benchmark for agentic CL research must satisfy two core requirements:
1) \textbf{Realistic task relations:} continual learning typically unfolds around the work process within a domain, so a benchmark should simulate how knowledge is reused in a task stream, rather than chaining arbitrary and isolated tasks into a sequence.
2) \textbf{Metrics for CL properties:} the evaluation protocol must decouple and quantify the properties of plasticity, stability, and generalization within the learning process beyond simply reporting average accuracy.

While more benchmarks focus on long-term agentic interactions, they fundamentally struggle to satisfy these requirements simultaneously.
First, long-term memory benchmarks \citep{locomo,longmemeval,memoryagentbench} evaluate whether agents can retrieve and reason over extended documents or conversations, yet they rely on static corpora rather than dynamic task adaptation.
Second, recent lifelong or streaming benchmarks \citep{streambench,lifelongagentbench,evo-memory} move closer to sequential adaptation, but they typically stream (i.i.d.) tasks from existing datasets, leaving cross-task relations unknown. 
It remains ambiguous whether an agent genuinely reused an abstracted experience or merely benefited from incidental task overlap.
Thus, they simply lack the scenarios to evaluate agentic experience reuse.
Besides, metric design is essential for evaluation, while existing benchmarks often rely on average accuracy, leaving CL properties such as plasticity and stability largely overlooked.

To bridge these fundamental gaps, we introduce \oureval, an evaluation framework for CL in language agents with \textbf{task streams} and \textbf{metrics}.
As shown in Figure~\ref{fig:eval}, \oureval distinguishes three cross-task relations. 
\textit{Conventional streams} leave cross-task reuse unknown, as in i.i.d.-built benchmarks.
Our proposed \textit{dependent streams} simulate capability-centered progression, where later tasks can reuse knowledge from earlier ones. \textit{Independent tasks} are held out from a different source, with little reusable knowledge.
Coupled with a two-pass evaluation protocol, \oureval allows us to separate and compute explicit metrics for \textit{Plasticity Gain}, \textit{Stability Gain}, and \textit{Generalization Gain}.

We leverage \oureval to conduct a comprehensive evaluation of non-parametric memory designs, a paradigm for inference-time learning without costly training.
We further develop \ourmethod, a non-parametric memory probing method for empirical studies.
Our empirical analysis across coding, deep research, and language reasoning tasks reveals a critical benchmarking insight: conventional streams offer surprisingly weak discriminative power, compressing the performance differences among memory designs. 
Conversely, our dependent streams clearly make experience reuse traceable and show varying plasticity across methods.
Moreover, our evaluation uncovers an unresolved stability bottleneck: while existing memory designs achieve notable transfer gains when reusable knowledge is present, they frequently induce memory interference or performance degradation on conventional streams and independent tasks.
These findings underscore the urgent need for memory designs that transcend passive retrieval to achieve robust, balanced consolidation.

Our core contributions are threefold:
1) We formalize \oureval, the first evaluation framework that explicitly distinguishes cross-task relations alongside metrics for plasticity, stability, and generalization.
2) We construct dependent streams, conventional streams, and independent held-out tasks spanning coding, deep research, and reasoning domains to capture whether reusable knowledge is available.
3) Comprehensive evaluation shows that dependent streams substantially improve the discriminative power of evaluation, while exposing critical stability bottlenecks in existing memory designs.

\section{Related Work on Continual Learning in Language Agents}
\label{sec:related_work}

\subsection{Benchmarks}
\label{subsec:benchmark-for-continual-learning}

The streaming setting has been studied in prior evaluation work.
StreamBench \citep{streambench} evaluates whether language agents can improve over an input--feedback sequence, while MemoryBench \citep{memorybench} focuses on whether LLM systems can construct useful memories from user feedback during service time. 
LifelongAgentBench \citep{lifelongagentbench} and SWE-Bench-CL \citep{swe-bench-cl} instantiate this idea in interactive agent and software-engineering domains, respectively, by organizing tasks into sequential settings that measure knowledge accumulation, transfer, and forgetting. 
Evo-Memory \citep{evo-memory} extends this direction by benchmarking self-evolving memory across diverse task streams and comparing multiple memory designs. 
Concurrent with our work, Continual Learning Bench \citep{continual-learning-bench} similarly evaluates agents over multi-episode environments with task relevance, such as recurring database schemas, debugging patterns, and forecasting dynamics.
While these benchmarks adopt online or streaming evaluation protocols, \oureval differs by explicitly contrasting different cross-task relation structures and by assuming no detailed user feedback for CL.

\subsection{Methodology}

Non-parametric memory records contexts in an intuitive way, but its applications to long-context or streaming tasks have not been rigorously evaluated for CL yet. 
Systems like Mem0 \citep{mem0} and A-MEM \citep{amem} focus on storing and organizing long-term memories for \textbf{adaptive retrieval}. 
Another line emphasizes \textbf{procedural reuse}: Agent Workflow Memory \citep{awm} induces reusable workflows from prior trajectories, and Dynamic Cheatsheet \citep{dc} maintains concise strategies, code snippets, and abstractions that can be reused at test time. 
ExpRAG and Evo-Memory's ReMem \citep{evo-memory} accumulate task experiences online and retrieve them in later tasks under \textbf{streaming} evaluation. 
Our evaluation targets not only how to store and retrieve experience, but also studies the learning dynamics from streams when experience reuse is available and when it is not.

Beyond non-parametric memory, recent work studies CL via parameter or policy updates in agentic settings. 
Agent-Dice \citep{agentdice} addresses stability--plasticity by disentangling shared and conflicting knowledge during parameter fusion for GUI and tool-use agents. 
CGL \citep{cgl} explores continual GUI learning with supervised and reinforcement fine-tuning to balance new-domain adaptation and capability retention. 
ACuRL \citep{acurl} autonomously explores environments, synthesizes capability-matched tasks, and updates policies across environments. 
These training-based methods internalize reusable experience into model parameters, but often require reward models, rollouts, and repeated optimization. 
\textbf{\textit{We instead focus on non-parametric memory as a lightweight inference-time adaptation. Evaluating training-based approaches with \oureval is a direction for future work.}}

\section{\oureval: Evaluation Framework for Continual Learning}
\label{sec:evaluation-framework}

\subsection{Problem Formulation}
\label{subsec:problem-formulation}

An agent observes a sequence of tasks $\mathcal{T}=\{\tau_1, \tau_2, \dots, \tau_T\}$ and solves them sequentially through interaction with the environment. 
In contrast to maintaining long contexts for each task, the agent maintains a persistent memory $\{M_t\}_{t=0}^T$ shared across the task stream $\mathcal{T}$. 
For each task $\tau_t$, the agent may first retrieve relevant information $c_t = \mathcal{R}(M_{t-1}, \tau_t)$ from the current memory, and then interact with the environment conditioned on $c_t$. 
After completing the task, it updates the memory via $M_t = \mathcal{U}(M_{t-1}, \xi_t)$, where $\xi_t$ denotes the trajectory collected during $\tau_t$.
Given the sparsity of feedback in real-world online environments, we adopt a strict setting where the evaluated agent operates without any ground-truth guidance during memory construction, leaving the LLM to synthesize memories entirely on its own.
In contrast to a memoryless setting, the order of tasks matters because the agent's accessible experience evolves over time.
Thus, we will proceed to discuss two core issues in benchmarking: the design of task streams and metrics.

\subsection{Task Relations and Streams}

Task relations and ordering determine what can be learned and evaluated.
Conventional task streams fail to capture a key property of real-world work: noise coexists with reusable experience.
We therefore distinguish three cross-task relations by the availability of reusable knowledge.
The first two are instantiated as task streams, on which memory is both built and evaluated. The third consists of held-out tasks evaluated with the memory frozen after either stream.
1) \textbf{Conventional:} tasks come from the same environment or domain, but whether reusable knowledge exists between them is unknown. This is the typical state of benchmarks built under an i.i.d. setting, which does not reflect continual learning around a capability within a domain.
2) \textbf{Dependent:} the stream simulates such capability-centered progression, where later tasks can reuse knowledge from earlier ones. These dependencies are hidden from the agent, which is never told which earlier task is relevant. Unrelated task pairs act as distractors for each other.
3) \textbf{Independent:} evaluated tasks are drawn from a different source than the memory-building stream, so that little transferable knowledge is available by design.

\subsection{Metrics}
\label{subsec:metrics}

\begin{figure}
    \centering
    \includegraphics[width=\linewidth]{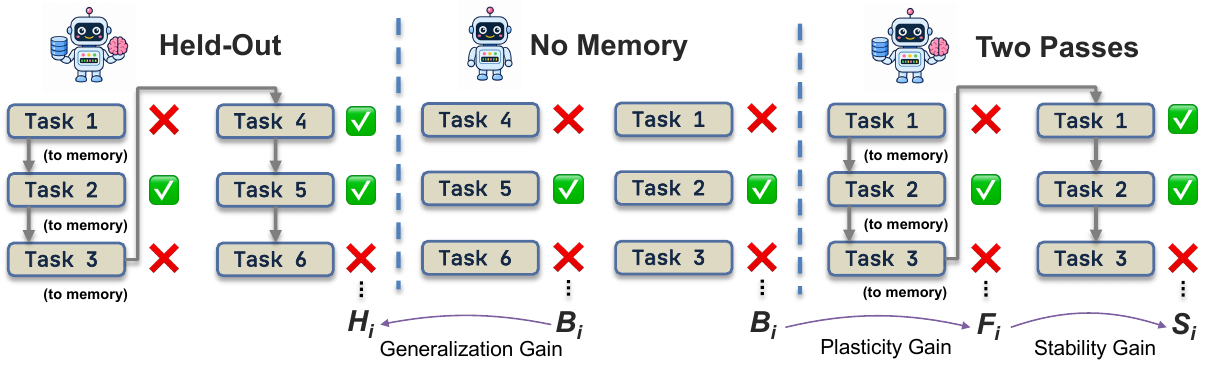}
    \caption{
Evaluation protocol and metrics of \oureval. 
\textbf{Middle}: memoryless baseline.
\textbf{Right}: two passes are performed on the same stream. 
The first pass allows both reading from and writing to memory, whereas the second pass (or independent evaluation) is read-only.
\textbf{Left}: independent tasks are not used to build memory.
For task $i$, $B_i$, $F_i$, $S_i$, and $H_i$ denote the performance of the memoryless method, the first-pass performance, the second-pass performance, and the performance on the independent task, respectively.
Plasticity Gain measures whether memory built from earlier tasks helps on later tasks.
Stability Gain measures whether experience gained from solving a task becomes persistently reusable.
Generalization Gain measures transfer to independent tasks after memory construction.
    }
    \label{fig:metrics}
\end{figure}

We first report basic metrics, such as accuracy, success rate, or progress rate, depending on the criterion of each environment.
Then, we focus on CL properties: \textit{plasticity}, \textit{stability}, and \textit{generalization} \citep{cl_survey}, as shown in Figure~\ref{fig:metrics}.
These properties from classical and parametric CL manifest differently in agentic settings.
Traditionally, stability is often framed as avoiding catastrophic forgetting after new tasks are absorbed into model weights. 
In contrast, we do not force an agent to learn through training, but mainly focus on non-parametric memory as the approach.
In this case, past experiences can be explicitly preserved as trajectories, summaries, skills, or other records. 
The central stability challenge is therefore not whether past episodes are recorded, but whether they can still be retrieved from memory and utilized correctly after many later updates.

This distinction from parametric CL motivates our two-pass evaluation protocol. 
The agent processes a task stream sequentially. 
The \textbf{first pass} permits both memory retrieval and updating, as described in \S\ref{subsec:problem-formulation}. 
The \textbf{second pass} is read-only with frozen memory.
Let \(B_i\) denote the performance on task \(i\) under a memoryless setting. 
Let \(F_i\) denote the performance on task \(i\) during the first pass, after the agent has processed tasks $1$ to $i-1$. 
Let \(S_i\) denote the performance on the same task $i$ during the second pass with frozen memory. 
We define
\begin{align}
  PG_i &= F_i - B_i, \\
  SG_i &= S_i - F_i.  
\end{align}

\textbf{Plasticity Gain} (\textbf{PG}) measures whether experience accumulated from preceding tasks helps the agent solve the current task. 
It captures the ability of a memory system to turn earlier episodes into useful context for later tasks.

\textbf{Stability Gain} (\textbf{SG}) measures persistent reuse under stream-level memory consolidation. 
As \(F_i\) is observed \textit{before} the experience from task \(i\) is written into memory, SG captures the net effect of writing the current task experience into memory, preserving it through subsequent updates, and retrieving it from the final frozen memory state. 
Thus, SG measures whether stream-level consolidation turns a task experience into stable support that remains accessible after later episodes are added. 
Low or negative SG reflects a failure of stable reuse, where the relevant experience is either poorly consolidated, difficult to retrieve, or overwhelmed by memory interference.

We further use independent tasks to evaluate generalization. 
Let \(H_j\) denote the performance on an independent task \(j\), evaluated after constructing and freezing memory on a task stream. 
We define
\begin{align}
  GG_j &= H_j - B_j.
\end{align}
\textbf{Generalization Gain} (\textbf{GG}) measures whether the constructed memory generalizes beyond the observed tasks used to build the memory.
Together, \oureval combines the designed task streams and metrics, as shown in Figure~\ref{fig:metrics}. 
Both conventional and dependent streams are used for two-pass evaluation. 
Different passes of a stream use the same tasks and sequence.

\section{Benchmark Construction}
\label{sec:benchmark-construction}

\subsection{Data Collection}
\label{subsec:data-collection}

\begin{table}[tb]
\small
\centering
\caption{The number of tasks in our constructed streams.
* denotes our synthesized data.
}
\label{tab:benchmark-instances}
\begin{tabularx}{\linewidth}{@{}Xll@{}}
\toprule
\textbf{Datasets} & \textbf{Dependent Stream} & \textbf{Conventional Stream} \\
\midrule
CodeEval-Pro (BigCodeBench-Lite-Pro) \citep{codeeval-pro}
& $48$ subtasks $\to$ $48$ complex tasks
& $48$ complex tasks \\

BrowseComp+ \citep{bc+}
& $308$ subtasks* $\to$ $100$ complex tasks 
& $100$ complex tasks \\

AgentBoard BabyAI \citep{agentboard}
& $28$ subtasks $\to$ $28$ complex tasks
& $40$ tasks \\

AgentBoard ScienceWorld \citep{agentboard}
& -- 
& $90$ tasks \\

MMLU-Pro \citep{mmlu_pro}
& -- 
& $300$ tasks \\
\bottomrule
\end{tabularx}
\end{table}

\noindent \textbf{Stream Design.} Table~\ref{tab:benchmark-instances} summarizes the constructed streams.
In real work within a domain, solutions to earlier, simpler tasks can often be reused in later, more complex ones; however, such reusable experience is sparse among many unrelated tasks, and the relations between tasks are usually implicit.
We construct each dependent stream by placing randomly ordered subtasks before the complex tasks of its conventional counterpart, and keep the subtask--task relations hidden from the agent.
As a result, each complex task can reuse the solution to an earlier subtask, yet on CodeEval-Pro and BrowseComp+ only about $1\%$ of the stored episodes are relevant when it arrives, and these episodes lie about half a stream earlier, so the agent must identify reusable experience among many distractors.
Since both streams evaluate exactly the same complex tasks, performance differences are attributable to the availability of reusable experience.
The construction of specific benchmarks across environments is detailed below.

\noindent \textbf{Coding.}
We build coding streams from CodeEval-Pro self-invoking problem pairs \citep{codeeval-pro}, which contain both BigCodeBench-Lite-Pro and HumanEval-Pro.
Each pair consists of a base problem (subtask) and a self-invoking problem (complex task) whose solution requires the solution to the base problem. 
Each task pair, from subtask to complex task, is guaranteed to exhibit a function reuse relation, as shown in Appendix Figure \ref{fig:codeeval-pro-example}.
Our filtered dataset includes $96$ tasks ($48$ pairs) from BigCodeBench-Lite-Pro.
The conventional stream randomly orders all complex tasks, which are diverse: their solutions may share a few Python packages, but their underlying logic differs without clear reuse relations.
We use $120$ tasks from HumanEval-Pro as independent tasks in \S\ref{subsec:discussion}.

\noindent \textbf{Deep Research.}
We build deep research streams based on BrowseComp+ \citep{bc+}. 
BrowseComp+ provides an offline environment with a collected corpus from BrowseComp \citep{bc}. 
We construct the streams similarly: the conventional stream contains $100$ randomly sampled BrowseComp+ tasks, while the dependent stream prepends $308$ subtasks. The subtasks are synthesized from these $100$ parent tasks by GPT-5.2 \citep{gpt52}.
The synthesis takes the original QA pair and the supporting documents as input, and yields subtasks and their answers.
Each subtask shares a subset of evidence documents with its parent complex task.
We set the reuse relations between subtasks and their corresponding complex tasks without hinting at the answers, as illustrated by the examples in Appendix~\ref{subsec:examples-of-compositionality} and details in Appendix~\ref{sec:new-assets}.

\noindent \textbf{Language Understanding \& Reasoning.}
To analyze streams from existing datasets, we select the single-turn MMLU-Pro and the multi-turn AgentBoard BabyAI and ScienceWorld tasks from Evo-Memory \citep{evo-memory}.
For MMLU-Pro \citep{mmlu_pro}, its tested knowledge is widely distributed, and we take it as a conventional stream.
We selected three subsets (economics, engineering, and philosophy) and sampled $100$ tasks from each subset.
For AgentBoard ScienceWorld \citep{agentboard}, we take all the $90$ original tasks as a conventional stream, and also evaluate a block stream in Appendix~\ref{subsec:agentboard-block}.
For AgentBoard BabyAI \citep{agentboard}, we build two streams, detailed in Appendix~\ref{subsec:details-in-benchmark-construction}. 
The conventional stream contains $40$ tasks from ten diverse easy subtask families, with four tasks per family.
The dependent stream contains $56$ tasks, where each task pair links a simpler subtask to a related compositional extension.

\subsection{Quality Control}

Most tasks used in our benchmark are directly adopted and restructured from existing resources.
For CodeEval-Pro, we apply simple filtering during restructuring. 
For synthesized BrowseComp+ subtasks, we apply stricter quality control.
Each candidate subtask must be evidence-supported, deterministic, and logically relevant to the parent task, and is further checked by a separate verifier for entity specificity, evidence support, determinism, and answer completeness. 
We also report diagnostics in Appendix~\ref{subsec:statistics-of-compositional-streams} to confirm that the subtasks do not collapse into parent-task paraphrases: the mean token-set Jaccard similarity between subtasks and their parent is $0.10$, no pair exceeds $0.30$, only $5.8\%$ of subtasks cover the full parent evidence set, and only $1.6\%$ share both the normalized answer and the evidence set with the parent task. 
These controls support compositional reuse through a chain of evidence documents rather than direct answer exposure.

\section{Experiments}
\label{sec:experiments}

\subsection{Setup}
\label{subsec:experiments-setup}

We evaluate representative methods from four categories. 
1) \textbf{Memoryless} agent denoted as ReAct \citep{react}. 
2) \textbf{Adaptive memory} methods that flexibly determine storage and retrieval: LangMem \citep{langmem} and Mem0 \citep{mem0}. 
3) Memory methods with \textbf{procedural knowledge}: Agent Workflow Memory (AWM) \citep{awm} and Dynamic Cheatsheet (DC-RS) \citep{dc}. 
4) \textbf{Self-evolving memory}: ExpRAG, Evo-Memory's ReMem \citep{evo-memory}, and \ourmethod. 
These categories span the main non-parametric designs that can approach CL in language agents.
We provide a detailed discussion of their coverage and differences in Appendix~\ref{subsec:memory-design-coverage}.

Given the tool support, we use Qwen3.5-35B-A3B \citep{qwen35a3b} with the stateful Qwen-Agent code interpreter by default, and gpt-oss-120b \citep{gptoss} with its natively trained browser tool for deep research.
Qwen3-Embedding-8B \citep{qwen3embedding} serves as the embedding model.
Experiments with other backbones show similar trends (Appendix~\ref{subsec:backbone-robustness}).
For cases where public code was unavailable or not directly applicable to our tasks, we reimplemented the method and matched the original workflow and prompts as closely as possible.
We report averages over three runs for coding tasks.
For all memory methods, we uniformly retrieve the top-$2$ memory chunks to serve as context for the current task, with more discussion in Appendix~\ref{subsec:retrieval-budget}.

\subsection{\ourmethod: Probing Memory Design}
\label{subsec:instrumented-method}

To diagnose how different components in memory designs affect CL, we introduce a non-parametric memory probing method, \ourmethod, which analyzes how agents accumulate and consolidate experience across a task stream while balancing plasticity and stability.
It adopts a retrieve--solve--consolidate loop over the task stream.
Before solving a task, it retrieves semantically similar prior episodes as the reference context.
After solving, it validates the agent responses and consolidates them into three complementary views: \textit{interaction} memory for concrete trajectories and final responses, \textit{insight} memory for task patterns and failure modes, and \textit{skill} memory for procedure-level abstractions or short reusable snippets.
Invalid outputs are discarded by a syntactic checker, and unreliable interaction histories are filtered by an LLM judge before being written into memory.
The prompt formats for constructing and using memory contexts are provided in Appendix~\ref{subsec:prompts}.
The procedure of \ourmethod is detailed in Appendix~\ref{subsec:algorithm}.

\subsection{Discussion}
\label{subsec:discussion}

\begin{figure}[tb]
    \centering
    \includegraphics[width=\linewidth]{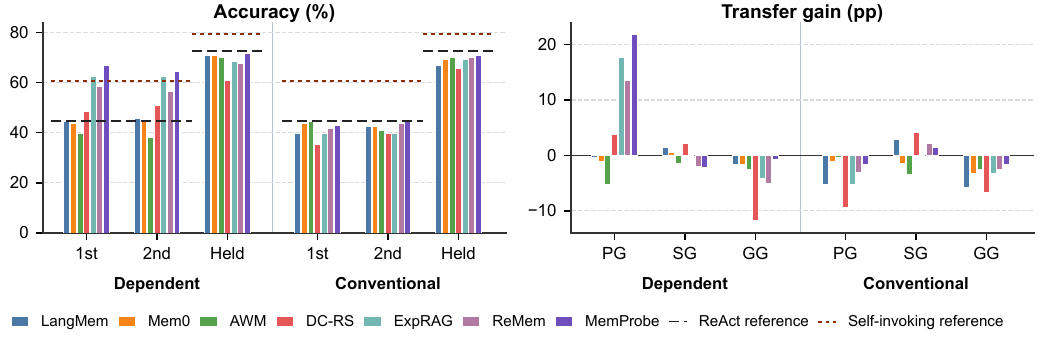}
    \caption{
Performance on CodeEval-Pro complex tasks. 
The BigCodeBench-Lite-Pro subset is used for conventional and dependent task streams. 
The HumanEval-Pro subset is used as independent tasks after the dependent or conventional stream. 
ReAct denotes a memoryless reference.
For self-invoking, each complex task is provided with its corresponding subtask as a reference context, so no retrieval is required.
Metrics are shown in Figure \ref{fig:metrics}.
Each value is the average over three runs. \textit{transfer gain} refers collectively to PG, SG, and GG, and \textit{pp} denotes percentage points.
    }
    \label{fig:codeeval-pro}
\end{figure}

\begin{figure}[tb]
    \centering
    \includegraphics[width=0.9\linewidth]{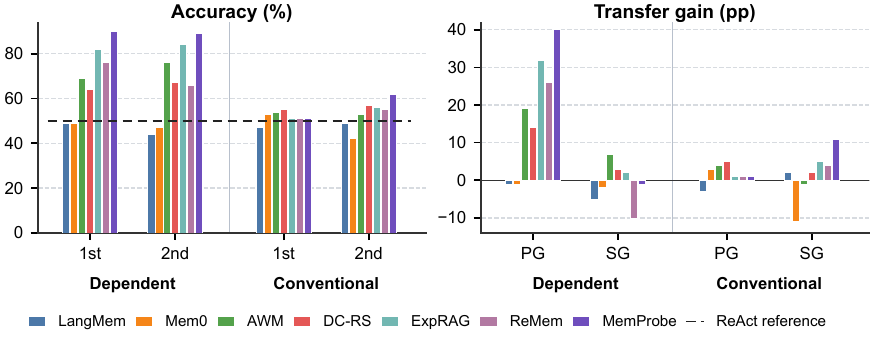}
    \caption{Performance on the selected $100$ original BrowseComp+ tasks.}
    \label{fig:bc+}
\end{figure}

\noindent \textbf{Dependent Streams Reveal Specialized Reuse.}
We show that dependent streams yield substantially clearer separation among memory methods than conventional streams. 
On CodeEval-Pro in Figure \ref{fig:codeeval-pro} and Table \ref{tab:bigcodebench}, the standard deviation of complex-task accuracy across methods is $9.4$ and $8.8$ in the two dependent-stream passes, compared with only $3.0$ and $1.9$ in the conventional stream. 
The same pattern is even more pronounced on BrowseComp+, where the corresponding dispersions are $14.9$ and $16.0$ under the dependent stream, but only $2.3$ and $5.7$ under the conventional stream, as shown in Figure \ref{fig:bc+} and Table \ref{tab:browsecomp}.
BabyAI is consistent with this trend in Table \ref{tab:babyai}.
By contrast, MMLU-Pro, evaluated as a conventional stream in Table \ref{tab:mmlu_pro}, shows only modest method separation, with average-accuracy standard deviations of $1.7$ and $1.8$ across the two passes.
These results suggest that designing task relations is essential to distinguish how different CL methods behave when reusable experience is available.
This pattern is further supported by pairwise bootstrap, task-distance, and stream-position analyses (Appendices~\ref{subsec:statistical-significance}, \ref{subsec:task-distance}, and~\ref{subsec:stream-position}), suggesting that the gains of existing memory designs primarily reflect specialized reuse.

\noindent \textbf{Conventional Streams Are Useful for Stability Tests.}
The unknown task relation in conventional streams is reflected in the first-pass results: in the conventional CodeEval-Pro stream in Figure \ref{fig:codeeval-pro}, PGs are compressed and mostly non-positive across memory methods.
MMLU-Pro also shows modest method separation across the two passes in Table \ref{tab:mmlu_pro}.
Conventional streams provide limited evidence about experience reuse, but they are still informative for stability evaluation: whether accumulated memory remains helpful, or at least harmless, after being updated by a stream with unknown reuse structure. 
In the conventional CodeEval-Pro stream, Appendix Table~\ref{tab:bigcodebench} shows that SG ranges from $-3.5$ to $+4.2$, indicating that consolidation helps some methods but hurts others.
BrowseComp+ in Appendix Table~\ref{tab:browsecomp} exposes even larger SG swings in the conventional stream, from $-11.0$ to $+11.0$. 
Appendix Table \ref{tab:scienceworld} further shows that even grouping ScienceWorld tasks by near-identical goal families produces limited method separation, suggesting that repeated exposure alone is not sufficient to create a diagnostic CL stream.

\noindent \textbf{Independent Tasks Are Necessary for Generalization Testing.}
Validation on held-out data is a standard practice in machine learning, yet not all the benchmarks for CL include such evaluation.
We use $120$ HumanEval-Pro tasks as independent tasks after the dependent or conventional stream from BigCodeBench-Lite-Pro.
They are still coding tasks but drawn from a different source and are not paired with any previous task.
As shown in Appendix Table~\ref{tab:bigcodebench}, GGs are limited for all memory methods, and memoryless ReAct remains strongest overall at $72.5$. 
Most memory methods perform slightly below ReAct, suggesting that memory accumulated from weakly related tasks does not automatically transfer and may introduce mild interference. 
The dependent and conventional streams provide different numbers of learning tasks, but the GGs after these two stream types are similar, indicating that the structural mismatch between learning and testing tasks is the main challenge, rather than the sheer number of preceding tasks. 
Among memory methods, \ourmethod is closest to ReAct at $71.7$ after the dependent stream and $70.8$ after the conventional stream, because it filters out irrelevant memory contexts in most cases. 
We additionally evaluate generalization on independent AgentBoard BabyAI tasks, with results reported in Appendix Table~\ref{tab:babyai-heldout}.

\noindent \textbf{Plasticity--Stability Tradeoff.}
These results expose a plasticity--stability tension in CL: methods that gain substantially when reusable knowledge is present are not necessarily stable across passes or stream types.
On CodeEval-Pro (Appendix Table~\ref{tab:bigcodebench}), ExpRAG, Evo-Memory's ReMem, and \ourmethod reach PG of $+17.7$, $+13.5$, and $+21.9$, yet their SG is only $+0.0$, $-2.0$, and $-2.1$, and they still trail memoryless ReAct on independent tasks (GG of $-4.2$, $-5.0$, and $-0.8$).
DC-RS attains the largest conventional-stream SG ($+4.2$) but suffers the largest GG drop ($-11.7$), suggesting that memory useful within a stream can hurt under distribution shift.
On BrowseComp+ (Appendix Table~\ref{tab:browsecomp}), ReMem gains $+26.0$ dependent-stream PG but loses $10.0$ SG, and Mem0 has the worst conventional-stream SG ($-11.0$) despite a slightly positive PG, whereas MMLU-Pro (Table~\ref{tab:mmlu_pro}) shows much smaller changes.
Thus, CL in language agents is not solved by storing more trajectories or summaries. The central challenge is deciding which past experience to abstract, retrieve, or ignore.
This imbalance is further tied to the data distribution: when reusable experience is scarce, an agent may need to actively explore its environment to build a useful memory context \citep{self-evolving}.

\noindent \textbf{Insights on Memory Design.}
Controlled streams help isolate the effects of memory design choices.
Table~\ref{tab:ablation_bigcodebench} reports ablations of \ourmethod on BigCodeBench-Lite-Pro. 
During the memory construction process, we have the syntactic checker and the LLM judge determine whether the output of an interaction is worth recording (\S\ref{subsec:instrumented-method}).
Replacing this validation step with the oracle boolean feedback yields similar overall performance, suggesting that the memory construction does not require a perfect judge.
Removing interaction, insight, or skill memory weakens performance on dependent streams in most cases.
In contrast, these ablations have smaller and less consistent effects on conventional streams. 
This supports the view that memory representations cannot be fairly diagnosed without task streams that expose opportunities for reuse.
The results also suggest that different forms of experience serve complementary roles. 
Interaction memory can preserve concrete solving episodes, insight memory can distill task-specific lessons and failure modes, and skill memory can capture reusable procedural knowledge.

\begin{table}[t]
\centering
\small
\caption{Accuracy (\%) on the conventional task stream from MMLU-Pro. \textit{Eco.}: economics. \textit{Eng.}: engineering. \textit{Phil.}: philosophy. Each domain contains $100$ samples. Memory is maintained separately by domain. \textit{Std. Dev.} denotes the standard deviation of the average metrics across evaluated methods. 
The highest value in each column is \textbf{bolded}, and the second-highest value is \underline{underlined}. 
The same applies to the following tables.}
\label{tab:mmlu_pro}
\begin{tabular}{lcccc|cccc}
\toprule
 & \multicolumn{4}{c}{1st Pass} & \multicolumn{4}{c}{2nd Pass} \\
\cmidrule(lr){2-5} \cmidrule(lr){6-9}
\textbf{Method} & Avg (PG) & Eco. & Eng. & Phil. & Avg (SG) & Eco. & Eng. & Phil. \\
\midrule
ReAct
& $78.7$
& $\underline{90.0}$
& $66.0$
& $\underline{80.0}$
& $78.7$
& $\underline{90.0}$
& $66.0$
& $\underline{80.0}$ \\

ExpRAG
& $\underline{81.3}$\ann{$\underline{+2.6}$}
& $\mathbf{91.0}$
& $73.0$
& $\underline{80.0}$
& $\mathbf{83.0}$\ann{$\underline{+1.7}$}
& $\mathbf{91.0}$
& $\underline{78.0}$
& $\underline{80.0}$ \\

ReMem
& $79.0$\ann{$+0.3$}
& $86.0$
& $\underline{74.0}$
& $77.0$
& $\underline{81.7}$\ann{$\mathbf{+2.7}$}
& $87.0$
& $\mathbf{81.0}$
& $77.0$ \\

\ourmethod
& $\mathbf{82.7}$\ann{$\mathbf{+4.0}$}
& $\mathbf{91.0}$
& $\mathbf{76.0}$
& $\mathbf{81.0}$
& $\mathbf{83.0}$\ann{$+0.3$}
& $\mathbf{91.0}$
& $76.0$
& $\mathbf{82.0}$ \\ \midrule
\textit{Std. Dev.}
& $1.7$
& $2.1$
& $3.8$
& $1.5$
& $1.8$
& $1.6$
& $5.6$
& $1.8$ \\
\bottomrule
\end{tabular}
\end{table}

\subsection{Case Study}
\label{subsec:case-study}

Table~\ref{tab:case_study} in Appendix~\ref{appendix-subsec:case-study} provides two coding cases where retrieved memories help or hurt. 
In the first case, \ourmethod retrieves a single-pair version of the same computation and reuses the correct DataFrame construction inside a loop. 
In the second case, the retrieved task is topically similar but semantically incompatible: it encourages selecting numeric columns, while the target task requires rejecting any non-numeric column. 
These cases illustrate that the effect of memory depends both on retrieval and on how the agent uses the retrieved context.

\begin{table}[t]
\centering
\small
\setlength{\tabcolsep}{5pt}
\caption{
Performance (\%) on AgentBoard BabyAI. \textit{SR}: success rate. \textit{PR}: progress rate. 
The conventional stream in AgentBoard BabyAI exhibits a pattern that tasks appear sequentially by task family, with each family comprising four tasks that share the same goal but feature different environmental settings.
}
\label{tab:babyai}
\begin{tabular}{lcccc|cccc}
\toprule
 & \multicolumn{4}{c|}{Dependent Stream} & \multicolumn{4}{c}{Conventional Stream} \\
\cmidrule(lr){2-5} \cmidrule(lr){6-9}
\textbf{Method} & 1-SR & 1-PR & 2-SR & 2-PR & 1-SR & 1-PR & 2-SR & 2-PR \\
\midrule
ReAct \citep{react} 
& $50.0$ 
& $63.6$ 
& $50.0$ 
& $63.6$ 
& $\underline{57.5}$ 
& $72.9$ 
& $57.5$ 
& $72.9$ \\

ExpRAG \citep{evo-memory} 
& $37.5$ 
& $53.1$ 
& $\underline{58.9}$ 
& $\underline{73.3}$ 
& $\mathbf{67.5}$ 
& $76.7$ 
& $\underline{70.0}$ 
& $78.7$ \\

ReMem \citep{evo-memory} 
& $\underline{51.8}$ 
& $\mathbf{68.6}$ 
& $\mathbf{62.5}$ 
& $\mathbf{75.7}$ 
& $\mathbf{67.5}$ 
& $\underline{77.9}$ 
& $\underline{70.0}$ 
& $\underline{79.2}$ \\

\ourmethod 
& $\mathbf{55.4}$ 
& $\underline{65.7}$ 
& $53.6$ 
& $69.0$ 
& $\mathbf{67.5}$ 
& $\mathbf{80.0}$ 
& $\mathbf{72.5}$ 
& $\mathbf{80.4}$ \\

\midrule
\textit{Std. Dev.}
& $6.7$
& $5.8$
& $4.8$
& $4.6$
& $4.3$
& $2.6$
& $5.9$
& $2.9$ \\
\bottomrule
\end{tabular}

\vspace{8pt}

\centering
\small
\setlength{\tabcolsep}{5pt}
\caption{
Accuracy (\%) of \ourmethod ablations on BigCodeBench-Lite-Pro with two passes.
\textit{Oracle feedback} means we replace the syntactic check and the LLM judge during memory consolidation with oracle feedback.}
\label{tab:ablation_bigcodebench}
\begin{tabular}{l|cc|cc}
\toprule
& \multicolumn{2}{c|}{Dependent} & \multicolumn{2}{c}{Conventional} \\
\textbf{Variant} & 1st & 2nd & 1st & 2nd \\
\midrule
\ourmethod & $\mathbf{66.7}$ & $\mathbf{64.6}$ & $43.1$ & $\underline{44.5}$ \\
\ \ \ \ w/ oracle feedback & $\underline{64.6}$ & $\underline{60.4}$ & $\underline{43.8}$ & $43.8$ \\
\ \ \ \ w/o skill & $56.3$ & $59.0$ & $\underline{43.8}$ & $\mathbf{45.8}$ \\
\ \ \ \ w/o insight & $60.4$ & $\mathbf{64.6}$ & $35.4$ & $41.7$ \\
\ \ \ \ w/o interaction & $52.1$ & $50.0$ & $\mathbf{45.8}$ & $41.7$ \\
\bottomrule
\end{tabular}
\end{table}

\section{Conclusion}
\label{sec:conclusion}

Existing benchmarks either emphasize long-context reasoning or build task streams without considering cross-task relations, which makes rigorous evaluation for agent CL difficult.
We propose an evaluation framework \oureval with task streams and metrics, contrasting conventional streams with dependent streams where earlier experience can support later tasks, and testing generalization on independent held-out tasks.
Across diverse agentic settings, empirical analysis indicates that: 
1) Task-stream design is essential for distinguishing CL capabilities. 
2) Balancing plasticity and stability remains a formidable challenge for agent memory, particularly in scenarios where opportunities for cross-task reuse are unclear.
These findings suggest that future CL benchmarks should publish task streams more transparently and discuss the relations between tasks, rather than simply streaming existing data and focusing solely on average metrics.

\subsection*{Ethics statement}
\label{sec:ethics-statement}

This work builds an evaluation framework based on publicly released datasets. 
Each asset and its license is listed in Appendix~\ref{sec:license-of-existing-assets}. 
The newly released artifacts, i.e., synthesized BrowseComp+ subtasks together with derived streams, are documented in Appendix~\ref{sec:new-assets}, including the LLM-based generation procedure and the quality control process.
Our continual learning method could potentially be exploited to develop harmful agents with progressively stronger capabilities. 
We believe that mitigating this risk mainly depends on the LLM’s ability to reject harmful requests.

\subsection*{Reproducibility statement}
\label{sec:reproducibility-statement}

We have taken the following steps to make our results reproducible. 
Complete reproduction instructions, covering the serving stack, model versions, decoding parameters, and agent interaction budgets, are given in Appendix~\ref{subsec:reproduction-instructions}.
The hardware and compute requirements are reported in Appendix~\ref{subsec:compute-resources}. 
All prompts used for memory construction, memory update, and evaluation are provided verbatim in Appendix~\ref{subsec:prompts}. The construction of task streams, including the data sources, filtering criteria, and the two-stage quality-control procedure applied to generated subtasks, is detailed in Appendix~\ref{subsec:details-in-benchmark-construction} and Appendix~\ref{sec:new-assets}.

\subsection*{Acknowledgments}

The authors would thank colleagues from the OSU NLP group and Intuit AI Research for constructive feedback. 
This research was sponsored in part by NSF CAREER \#1942980, NSF CAREER \#2443149, the Alfred P. Sloan Research Fellowship, and gifts from Intuit Inc. 
The views and conclusions contained herein are those of the authors and should not be interpreted as representing the official policies, either expressed or implied, of the U.S. government.
The U.S. Government is authorized to reproduce and distribute reprints for Government purposes notwithstanding any copyright notice herein.

\bibliographystyle{plain}
\bibliography{ref}




\clearpage
\appendix
\AppendixTOC

\makeatletter
\setlength{\@fptop}{0pt}
\setlength{\@fpsep}{12pt plus 1fil}
\setlength{\@fpbot}{0pt plus 1fil}
\makeatother
\setcounter{topnumber}{4}
\setcounter{bottomnumber}{2}
\setcounter{totalnumber}{6}
\renewcommand{\floatpagefraction}{0.85}

\newpage
\section{Limitations}
\label{sec:limitations}

This paper focuses primarily on continual learning for language agents through non-parametric memory. 
This scope is appropriate for isolating how reusable experience can be accumulated and applied across task streams without changing model parameters. 
However, the current study does not systematically evaluate parametric memory or training-based adaptation methods, which may offer complementary advantages in continual learning. 
Incorporating such approaches into the evaluation framework would be a valuable direction for future work.

\section{Experimental Setting and Details}
\label{sec:appendix-experimental-settings}

\subsection{Examples of Compositionality}
\label{subsec:examples-of-compositionality}

\begin{figure}[tbh]
    \centering
    \includegraphics[width=\linewidth]{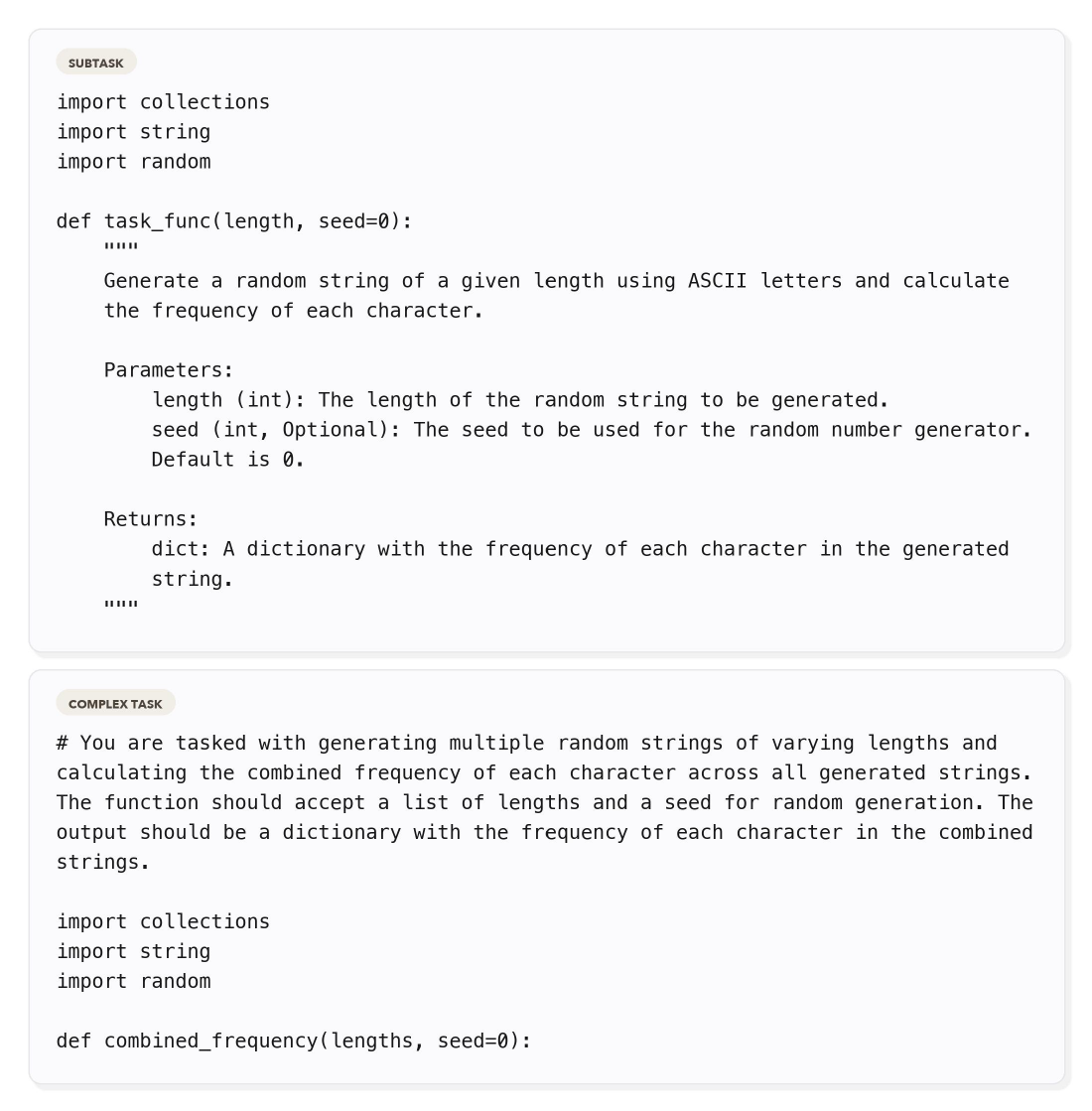}
    \caption{
    Example of a CodeEval-Pro task pair. The subtask computes character frequencies for one generated string, while the complex task extends the same logic to aggregate frequencies across multiple random strings of varying lengths.
}
    \label{fig:codeeval-pro-example}
\end{figure}

\begin{figure}[tbh]
    \centering
    \includegraphics[width=\linewidth]{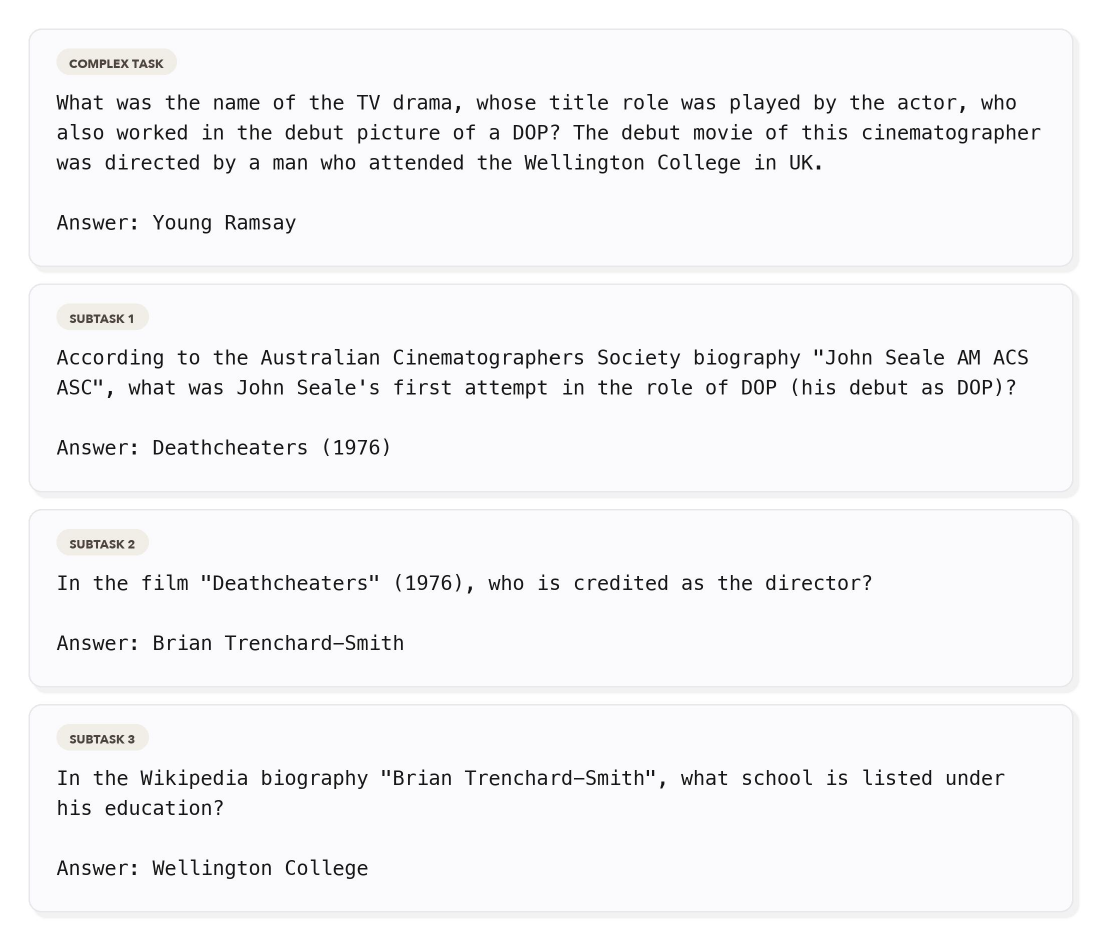}
    \caption{
    Example of a BrowseComp+ task and our synthesized subtasks. The subtasks derived from the original task are distinctly different. The answers are provided solely to aid the reader’s understanding. During the solving process, the agent has access only to the task statement and the corpus. The synthesis of these subtasks ensures that even if the agent reads all the subtask statements, it cannot directly derive the answer to the original task without research.
    }
    \label{fig:bc-plus-example}
\end{figure}

Figure~\ref{fig:codeeval-pro-example} illustrates an example task pair from CodeEval-Pro, where the complex task preserves the core solution logic of the subtask while requiring composition over multiple generated strings. Figure~\ref{fig:bc-plus-example} further shows an example from BrowseComp+, where our synthesized subtasks are not simple decompositions that directly reveal the original answer, but instead share intrinsically related solution logic with the original task. 
Together, these examples demonstrate how our construction exposes reusable problem-solving patterns across related tasks while avoiding direct answer leakage.

\subsection{Evaluated Memory Designs}
\label{subsec:memory-design-coverage}

Our evaluation is designed to cover the main non-parametric memory design choices that can support continual learning in language agents.
We organize representative methods by what they store, how they update memory, and how the memory is reused during later task solving.

\paragraph{Memoryless reference.}
We take ReAct \citep{react} as a memoryless reference method.
The tools available to the agent vary depending on the specific environment. Generally, we employ LLMs equipped with built-in reasoning capabilities, which means these models think before invoking a tool or responding to the user.
During each round of interaction with the environment, the preceding reasoning process is preserved. However, this reasoning process cannot be carried over to a subsequent, new task.

\paragraph{Self-invoking reference.}
We adopt the self-invoking method as a reference for CodeEval-Pro \citep{codeeval-pro}, as it provides an intuitive solution to self-invoking tasks that we take as a source for our task stream.
Note that self-invoking uses inputs distinct from our default experimental setting. It is not treated as a method for direct comparison, but serves as a reference. 
Each complex task is accompanied by corresponding subtasks as input without any other memory contexts, thereby assessing the LLM's capacity for self-invoking the sub-solution rather than its ability for continual learning.
A variation of the self-invoking method is to provide a solution to the subtask alongside its description. 
Although the input configurations differ, the two self-invoking variants constitute a meaningful comparison because memory methods do not uniformly fall above or below these references.

\paragraph{Adaptive semantic and episodic memory.}
LangMem \citep{langmem} and Mem0 \citep{mem0} represent adaptive long-term memory systems that decide what information should be extracted, consolidated, and retrieved.
This family mainly treats memory as persistent natural-language or structured records that can be searched and injected into future contexts.
LangMem provides general primitives for extracting semantic, episodic, and procedural memories from interactions, updating them with an LLM-based memory manager, and retrieving them through an external store.
Mem0 similarly builds a scalable memory layer that dynamically extracts salient information from ongoing interactions, compares new candidates with existing memories, and performs memory operations such as adding, updating, deleting, or ignoring entries.
These methods cover the design where memory is primarily an adaptive store of facts, events, or prior episodes, and the central question is whether a memory manager can preserve useful information while avoiding redundancy and conflict.

\paragraph{Procedural memory.}
AWM \citep{awm} and DC-RS \citep{dc} represent methods that focus on procedural knowledge.
AWM induces reusable workflows from successful agent trajectories.
A workflow abstracts away instance-specific details and stores a reusable routine that can guide future action generation, making it suitable for long-horizon interactive tasks where similar sub-procedures recur across tasks.
DC-RS maintains an evolving cheatsheet of strategies, heuristics, and code snippets.
It retrieves similar past input--output pairs and synthesizes them with the current memory to update the cheatsheet before solving the next task.
This family covers the design where memory is not merely episodic records, but a compact representation of how to solve recurring classes of problems.

\paragraph{Self-evolving memory.}
ExpRAG, ReMem \citep{evo-memory}, and \ourmethod represent self-evolving memory methods that explicitly operate over a task stream.
ExpRAG is a simple experience-retrieval method: it stores previous task experiences, retrieves similar ones for the current task, and uses them as in-context demonstrations.
This isolates the effect of experience-level retrieval without adding more complex memory reasoning.
ReMem goes further by integrating memory refinement into the agent loop, allowing the agent to reason, act, and refine memory as interacting components rather than treating memory as a passive retrieved context.
Our \ourmethod follows the same broad direction but is designed as a probing method for this benchmark.
It separates interaction, insight, and skill memories and applies quality-aware consolidation before writing experiences into memory.
Together, these methods cover the design where memory is expected to evolve during deployment and where the key challenge is not only retrieval, but also deciding which experiences should be abstracted, preserved, or ignored over time.

\subsection{Reproduction Instructions}
\label{subsec:reproduction-instructions}

We deployed the LLM and embedding model using vLLM v0.16.1.\footnote{\url{https://github.com/vllm-project/vllm}}
For the Qwen 3.5 model, we used thinking mode and adopted the sampling parameters recommended in the official documentation for precise coding tasks: temperature $= 0.6$, top\_p $= 0.95$, top\_k $= 20$, min\_p $= 0.0$, presence\_penalty $= 0.0$, repetition\_penalty$= 1.0$.\footnote{\url{https://huggingface.co/Qwen/Qwen3.5-35B-A3B}}
For the OpenAI gpt-oss model, we set reasoning effort to \textit{high} and used default values for the other hyperparameters in the OpenAI client.
All deployed LLMs and embedding models are official, non-quantized versions.
For hyperparameters, we set the maximum number of tool calls for BrowseComp+ to $100$, and the maximum number of action steps for AgentBoard environments to $50$. 
For memory methods, we used a shared top-$k$ retrieval setting with $k=2$.

\subsection{Compute Resources}
\label{subsec:compute-resources}

We use one NVIDIA A100 to deploy Qwen3-Embedding-8B, and at least two NVIDIA A100s to deploy Qwen3.5-35B-A3B or openai/gpt-oss-120b.
The required time depends on the number of deployed GPUs, the size of the dataset, and the number of inference steps. 
We report the time costs of \ourmethod in Table \ref{tab:time_cost}.
In terms of runtime and the number of tool calls, reusing experience reduces the cost of solving complex tasks.
We also report the token usage of evaluated methods on dependent streams in Table \ref{tab:compositional_token_usage}.

\begin{table}[tb]
    \centering
    \small
    \caption{Average execution statistics on complex tasks under conventional and dependent streams. \textit{Search calls} refer to the number of times the document corpus is retrieved, rather than the number of memory retrievals. Search calls are included in tool calls.}
    \label{tab:time_cost}
    \begin{tabular}{llrrrr}
        \toprule
        \textbf{Benchmark} & \textbf{Stream} & \textbf{Time (s)} & \textbf{Tool Calls} & \textbf{Search Calls} & \textbf{LLM Calls} \\
        \midrule
        \multirow{2}{*}{BigCodeBench-Lite-Pro}
            & Conventional        & $26.2$  & $1.9$  &    --   & $2.9$ \\
            & Dependent & $24.4$  & $1.7$  &    --  & $2.7$ \\
        \midrule
        \multirow{2}{*}{BrowseComp+}
            & Conventional        & $109.7$ & $49.9$ & $31.4$ & $52.8$ \\
            & Dependent & $46.2$  & $30.7$ & $16.2$ & $32.8$ \\
        \bottomrule
    \end{tabular}
\end{table}

\begin{table}[tb]
\centering
\small
\caption{Token usage statistics on the BigCodeBench-Lite-Pro and BrowseComp+ dependent streams ($96$ tasks and $408$ tasks, respectively; subtasks included). Both datasets are evaluated over two passes. All values are reported in millions of tokens. The tokenizers are derived from the corresponding LLMs.}
\label{tab:compositional_token_usage}
\begin{tabular}{lrrr|rrr}
\toprule
\textbf{Method}
& \multicolumn{3}{c|}{\textbf{BigCodeBench-Lite-Pro}}
& \multicolumn{3}{c}{\textbf{BrowseComp+}} \\
\cmidrule(lr){2-4}
\cmidrule(lr){5-7}
& \textbf{Total} & \textbf{Input} & \textbf{Output}
& \textbf{Total} & \textbf{Input} & \textbf{Output} \\
\midrule

LangMem
& $1.8$ & $1.4$ & $0.4$
& $542.5$ & $539.1$ & $3.4$ \\

Mem0
& $2.5$ & $2.1$ & $0.4$
& $543.6$ & $540.2$ & $3.4$ \\

AWM 
& $3.9$ & $3.4$ & $0.5$
& $375.9$ & $373.3$ & $2.7$ \\

DC-RS 
& $6.2$ & $5.7$ & $0.5$
& $412.7$ & $409.8$ & $2.9$ \\

ExpRAG 
& $1.9$ & $1.6$ & $0.3$
& $256.7$ & $254.5$ & $2.2$ \\

ReMem
& $1.6$ & $1.2$ & $0.3$
& $357.6$ & $355.0$ & $2.7$ \\

\ourmethod
& $2.5$ & $2.1$ & $0.4$
& $220.9$ & $218.9$ & $2.0$ \\
\bottomrule
\end{tabular}
\end{table}

\subsection{Prompts}
\label{subsec:prompts}

\begin{figure}[!htb]
    \centering
    \includegraphics[width=0.95\linewidth]{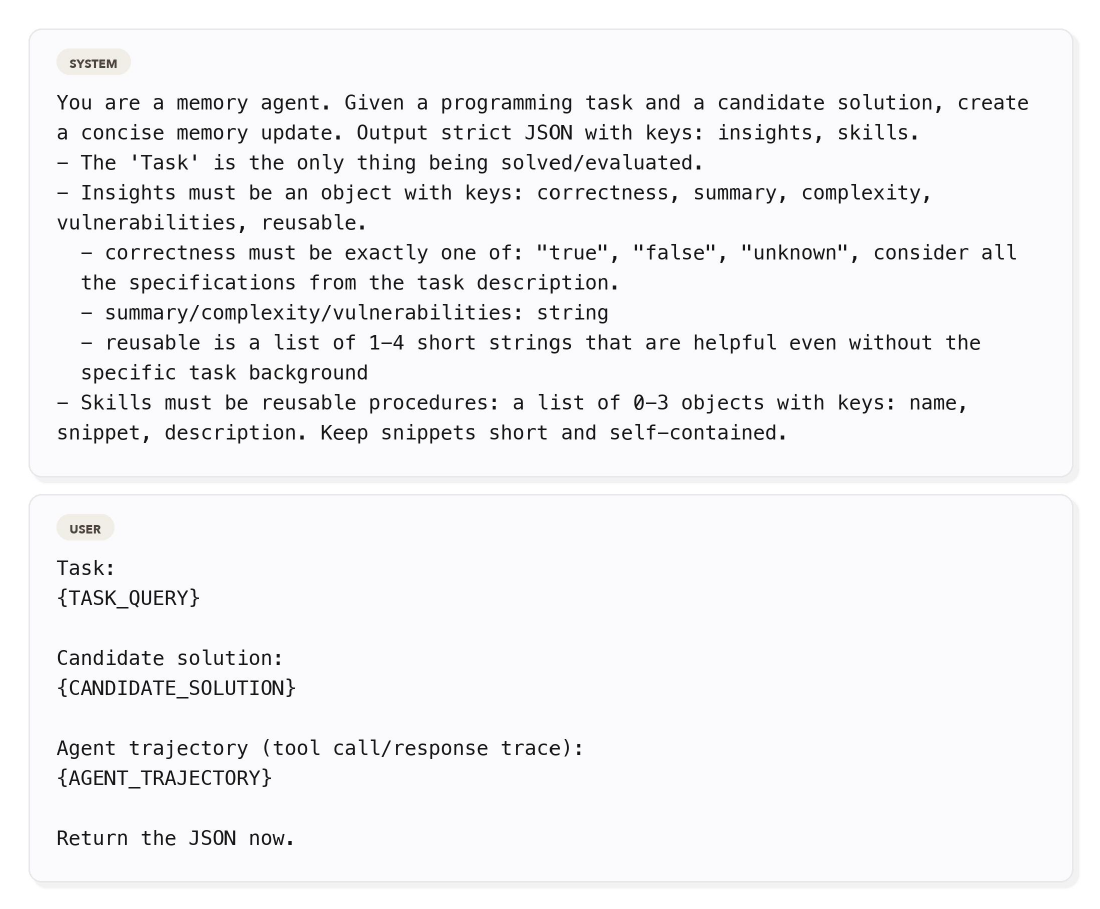}
    \caption{Memory update prompt in the coding setting. The interactive trajectory is stored as interaction memory, while an LLM extracts the current task episode into insight and skill memories.}
    \label{fig:memory-update-prompt}
\end{figure}

\begin{figure}[!htb]
    \centering
    \includegraphics[width=0.95\linewidth]{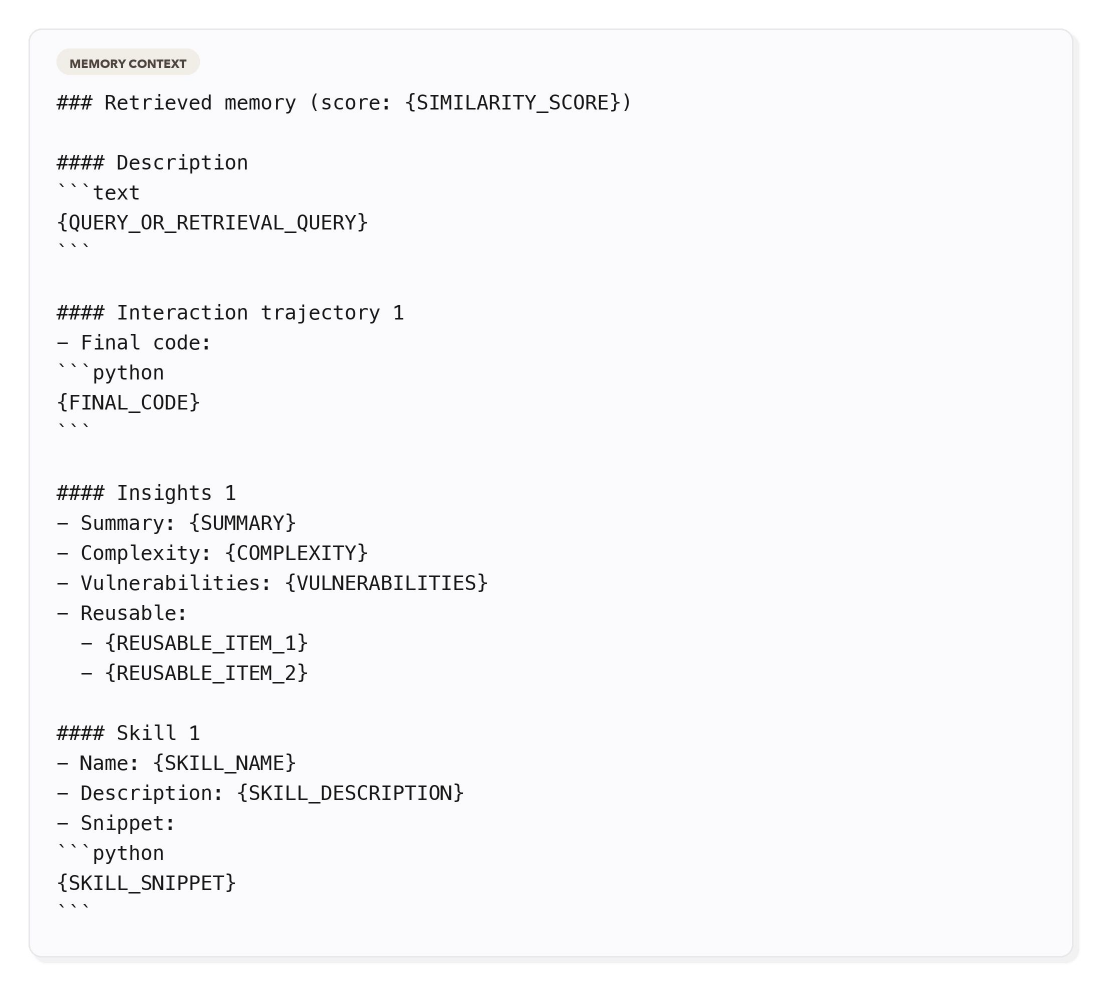}
    \caption{Retrieved memory context in the coding setting. For each retrieved task, the solver is provided with the task description, interaction memory, extracted insights, and reusable skills.}
    \label{fig:memory-context-template}
\end{figure}

Figure \ref{fig:memory-update-prompt} and Figure \ref{fig:memory-context-template} illustrate the prompting and rendering design used in \ourmethod under the coding setting. 
After each task, the memory update prompt in Figure \ref{fig:memory-update-prompt} takes the task, the candidate solution, and the agent’s generated code as input, and asks an LLM to extract the episode into structured insight and skill memories, while the trajectory, i.e., generated code in this task, is retained as interaction memory. 
At inference time, as shown in Figure \ref{fig:memory-context-template}, retrieved memories from prior tasks are organized into a unified memory context that includes the task description, interaction record, extracted insights, and reusable skills, which are then provided to the agent to support solving the current task. 

\subsection{Details of \ourmethod}
\label{subsec:algorithm}

To instantiate \oureval, we introduce a non-parametric memory probing method \ourmethod for analyzing how agents accumulate and consolidate experience across a task stream.
It aims for a balance between plasticity and stability.
For plasticity, memory should capture complementary forms of experience, such as raw trajectories, concise summaries, and reusable skills, rather than biasing the agent toward a single representation that may only work in specific domains.
For stability, stored experience should improve future decisions more often than it harms them, since past solutions can also bias the agent toward incorrect reasoning paths.
Thus, we introduce \ourmethod based on a retrieve--solve--consolidate loop.

\paragraph{Memory Structure.}

Rather than storing only raw trajectories or only compact summaries, \ourmethod organizes experience into three complementary memory types.
1) \textbf{Interaction memory}: This memory records the concrete solving episode, including the agent's interaction trajectory and the final response. It is useful when future tasks benefit from recalling how a similar problem was operationally solved.
2) \textbf{Insight memory}: This memory stores extracted information from the episode, including a concise summary of the task-solving pattern, complexity-related observations, potential failure modes, and reusable takeaways. Compared with interaction memory, it is more abstract and compact, making it suitable for transferring task-relevant lessons without replaying the detailed trajectory.
3) \textbf{Skill memory:} This memory stores reusable procedures or workflow-level knowledge in a more explicit form, optionally accompanied by short code snippets. Skill memory is intended to capture higher-level strategies that may generalize across multiple tasks, rather than being tied to one specific instance.
This decomposition reflects our view that reusable experience in continual learning is heterogeneous. 
Comprehensive representation allows the method to represent cross-task experience more flexibly than a single-memory format.

\paragraph{Memory Retrieval and Reuse.}

Given a new task $\tau_t$, \ourmethod first retrieves a small set of relevant past experiences from memory using embedding-based similarity search over previously stored task descriptions. Let $M_{t-1}$ denote the current memory state and let $\mathcal{R}$ denote retrieval. The retrieved context is
\[
c_t = \mathcal{R}(M_{t-1}, \tau_t),
\]
where retrieval returns the top-$k$ most similar prior tasks above a similarity threshold, together with the interaction, insight, and skill entries associated with those tasks.
In our implementation, retrieval operates at the task level: the current task is matched against previously stored task descriptions via semantic similarity, and the associated interaction, insight, and skill entries are then rendered into the memory context.
The agent then solves $\tau_t$ conditioned on the retrieved context. 
Importantly, retrieved memory is treated as reference context rather than authoritative guidance, so the model is encouraged to ground its response in the current task instead of reproducing superficially similar past solutions or trajectories.
A small retrieval budget bounds reuse by exposing only top-$k$ semantically related entries, so that interference from weakly related tasks is reduced.

\paragraph{Quality-Aware Memory Consolidation.}
\label{subsec:quality}

After completing task $\tau_t$, the method consolidates the current episode into a new memory record:
\[
M_t = \mathcal{U}(M_{t-1}, \tau_t, \xi_t).
\]
During consolidation, the episode is converted into the three memory types described above. The updated memory is immediately available to subsequent tasks, enabling online accumulation of experience over the task stream.
A central challenge in continual memory is that solved episodes vary in future utility. Poor solutions, malformed outputs, or misleading trajectories may be retrieved later and degrade decision-making. To mitigate this risk, \ourmethod incorporates quality-aware controls during consolidation.
First, consolidation is gated by basic solution validity. In the coding setting, the system skips consolidation if the extracted solution is not syntactically valid, preventing obviously unusable outputs from entering long-term memory.
Second, the consolidation model is prompted to judge whether the candidate solution is correct, incorrect, or uncertain for the current task. 
This signal is stored in the insight memory and used conservatively during future reuse. In particular, memories judged clearly incorrect are not allowed to contribute full interaction-level traces as reusable context. This preserves compact evidence that a past attempt failed, while reducing the risk of replaying erroneous reasoning in later tasks.
Overall, \ourmethod provides a conservative continual learning mechanism that steadily accumulates structured experience from multi-turn agent episodes, while carefully restricting how experience is consolidated and reused so that memory is more likely to aid transfer than amplify noise.

\paragraph{Algorithm.}
Algorithm~\ref{alg:continent} summarizes the online memory update and reuse process of \ourmethod over a task stream $\mathcal{T} = (\tau_1, \tau_2, \dots, \tau_T)$, where $\tau_t$ denotes the $t$-th task, $M_t$ denotes the memory state after finishing $\tau_t$, and $\xi_t$ denotes the interaction trajectory collected while solving $\tau_t$. 

\begin{algorithm}[t]
\caption{\ourmethod: Online Memory Update and Reuse}
\label{alg:continent}
\begin{algorithmic}[1]
\REQUIRE Task stream $\mathcal{T} = (\tau_1, \tau_2, \dots, \tau_T)$; retrieval function $\mathcal{R}$; agent policy $\pi$; validation function $\mathcal{V}$; extraction function $\mathcal{D}$; memory update function $\mathcal{U}$; retrieval budget $k$
\ENSURE Final memory $M_T$ and task outputs $\{a_t\}_{t=1}^T$

\STATE Initialize memory $M_0 \leftarrow \emptyset$
\FOR{$t = 1, \ldots, T$}
    \STATE Retrieve relevant prior experience:
    \[
    c_t \leftarrow \mathcal{R}(M_{t-1}, \tau_t, k)
    \]
    \STATE Solve the current task:
    \[
    (a_t, \xi_t) \leftarrow \pi(\tau_t, c_t)
    \]
    \STATE Validate the episode, where the quality assessment $q_t \in \{\text{correct}, \text{unknown}, \text{incorrect}\}$:
    \[
    q_t \leftarrow \mathcal{V}(\tau_t, a_t, \xi_t)
    \]
    \STATE Extract structured memories for interaction, insights, and skills:
    \[
    (m_t^{\text{int}}, m_t^{\text{ins}}, m_t^{\text{skill}})
    \leftarrow \mathcal{D}(\tau_t, a_t, \xi_t, q_t)
    \]
    \STATE Update persistent memory:
    \[
    M_t \leftarrow \mathcal{U}(M_{t-1}, m_t^{\text{int}}, m_t^{\text{ins}}, m_t^{\text{skill}})
    \]
\ENDFOR
\RETURN $M_T$ and $\{a_t\}_{t=1}^T$
\end{algorithmic}
\end{algorithm}

\subsection{Statistical Significance}
\label{subsec:statistical-significance}

We present mean values with $95\%$ bootstrap confidence intervals (CIs) for the main results in Tables~\ref{tab:bigcodebench_ci} and \ref{tab:browsecomp_ci}.

\begin{table}[tbp]
\centering
\small
\caption{Performance on CodeEval-Pro (BigCodeBench-Lite-Pro and HumanEval-Pro) with 95\% bootstrap CIs. 
Each value is the average accuracy over three runs, with CIs shown as superscript and subscript.}
\label{tab:bigcodebench_ci}
\setlength{\tabcolsep}{3pt}
\begin{tabular}{l|ccc|ccc}
\toprule
 & \multicolumn{3}{c|}{Dependent} & \multicolumn{3}{c}{Conventional} \\
\textbf{Pass} & 1st & 2nd & Indep. & 1st & 2nd & Indep. \\ 
\midrule
Self-invoking
& $60.4^{\scriptscriptstyle +8.3}_{\scriptscriptstyle -7.6}$ 
& $60.4^{\scriptscriptstyle +8.3}_{\scriptscriptstyle -7.6}$ 
& --
& -- & -- 
& $79.2^{\scriptscriptstyle +4.2}_{\scriptscriptstyle -4.2}$ \\

\ \ w/ subtask solution
& $59.0^{\scriptscriptstyle +8.3}_{\scriptscriptstyle -8.3}$ 
& $59.0^{\scriptscriptstyle +8.3}_{\scriptscriptstyle -8.3}$ 
& --
& -- & -- 
& $73.3^{\scriptscriptstyle +4.5}_{\scriptscriptstyle -4.7}$ \\
\midrule

ReAct
& $44.8^{\scriptscriptstyle +8.3}_{\scriptscriptstyle -7.6}$ 
& $44.8^{\scriptscriptstyle +8.3}_{\scriptscriptstyle -7.6}$ 
& $\mathbf{72.5}^{\scriptscriptstyle +4.7}_{\scriptscriptstyle -4.7}$
& $\mathbf{44.8}^{\scriptscriptstyle +8.3}_{\scriptscriptstyle -7.6}$ 
& $\mathbf{44.8}^{\scriptscriptstyle +8.3}_{\scriptscriptstyle -7.6}$ 
& $\mathbf{72.5}^{\scriptscriptstyle +4.7}_{\scriptscriptstyle -4.7}$ \\

LangMem
& $44.4^{\scriptscriptstyle +8.3}_{\scriptscriptstyle -8.3}$ 
& $45.8^{\scriptscriptstyle +8.3}_{\scriptscriptstyle -8.3}$ 
& {$70.8^{\scriptscriptstyle +4.7}_{\scriptscriptstyle -4.7}$}
& $39.6^{\scriptscriptstyle +8.3}_{\scriptscriptstyle -7.6}$ 
& $42.4^{\scriptscriptstyle +8.3}_{\scriptscriptstyle -7.6}$ 
& $66.7^{\scriptscriptstyle +4.7}_{\scriptscriptstyle -5.0}$ \\

Mem0
& $43.8^{\scriptscriptstyle +8.3}_{\scriptscriptstyle -8.3}$ 
& $44.4^{\scriptscriptstyle +8.3}_{\scriptscriptstyle -8.3}$ 
& {$70.8^{\scriptscriptstyle +4.7}_{\scriptscriptstyle -4.7}$}
& $43.8^{\scriptscriptstyle +8.3}_{\scriptscriptstyle -7.7}$ 
& $42.4^{\scriptscriptstyle +8.3}_{\scriptscriptstyle -7.6}$ 
& $69.2^{\scriptscriptstyle +4.7}_{\scriptscriptstyle -4.7}$ \\

AWM
& $39.6^{\scriptscriptstyle +7.6}_{\scriptscriptstyle -7.6}$ 
& $38.2^{\scriptscriptstyle +8.3}_{\scriptscriptstyle -7.6}$ 
& $70.0^{\scriptscriptstyle +4.7}_{\scriptscriptstyle -4.7}$
& $44.5^{\scriptscriptstyle +8.3}_{\scriptscriptstyle -8.3}$ 
& $41.0^{\scriptscriptstyle +8.3}_{\scriptscriptstyle -7.6}$ 
& $70.0^{\scriptscriptstyle +4.7}_{\scriptscriptstyle -4.7}$ \\

DC-RS
& $48.6^{\scriptscriptstyle +7.6}_{\scriptscriptstyle -8.3}$ 
& $50.7^{\scriptscriptstyle +8.3}_{\scriptscriptstyle -8.3}$ 
& $60.8^{\scriptscriptstyle +5.0}_{\scriptscriptstyle -5.0}$
& $35.4^{\scriptscriptstyle +7.6}_{\scriptscriptstyle -7.6}$ 
& $39.6^{\scriptscriptstyle +7.6}_{\scriptscriptstyle -7.6}$ 
& $65.8^{\scriptscriptstyle +5.0}_{\scriptscriptstyle -5.0}$ \\

ExpRAG
& $\underline{62.5}^{\scriptscriptstyle +7.6}_{\scriptscriptstyle -7.6}$ 
& $\underline{62.5}^{\scriptscriptstyle +7.6}_{\scriptscriptstyle -7.6}$ 
& $68.3^{\scriptscriptstyle +4.7}_{\scriptscriptstyle -4.7}$
& $39.6^{\scriptscriptstyle +7.6}_{\scriptscriptstyle -7.6}$ 
& $39.6^{\scriptscriptstyle +7.8}_{\scriptscriptstyle -7.8}$ 
& $69.2^{\scriptscriptstyle +4.7}_{\scriptscriptstyle -4.7}$ \\

ReMem
& $58.3^{\scriptscriptstyle +7.7}_{\scriptscriptstyle -8.3}$ 
& $56.3^{\scriptscriptstyle +8.3}_{\scriptscriptstyle -7.6}$ 
& $67.5^{\scriptscriptstyle +5.0}_{\scriptscriptstyle -5.0}$
& $41.7^{\scriptscriptstyle +8.3}_{\scriptscriptstyle -7.7}$ 
& $43.8^{\scriptscriptstyle +7.7}_{\scriptscriptstyle -8.3}$ 
& $70.0^{\scriptscriptstyle +4.7}_{\scriptscriptstyle -4.7}$ \\

\ourmethod 
& $\mathbf{66.7}^{\scriptscriptstyle +7.6}_{\scriptscriptstyle -7.6}$ 
& $\mathbf{64.6}^{\scriptscriptstyle +7.6}_{\scriptscriptstyle -8.3}$ 
& $\underline{71.7}^{\scriptscriptstyle +4.7}_{\scriptscriptstyle -4.7}$
& \underline{$43.1^{\scriptscriptstyle +7.6}_{\scriptscriptstyle -7.6}$} 
& \underline{$44.5^{\scriptscriptstyle +8.3}_{\scriptscriptstyle -7.6}$} 
& \underline{$70.8^{\scriptscriptstyle +4.7}_{\scriptscriptstyle -4.7}$} \\
\bottomrule
\end{tabular}
\end{table}

\begin{table}[tbp]
\centering
\small
\caption{Performance on BrowseComp+ ($308$ subtasks and $100$ tasks). \textit{Sub} denotes subtasks, and \textit{Q} denotes complex tasks.}
\label{tab:browsecomp_ci}
\setlength{\tabcolsep}{3pt}
\begin{tabular}{l|cccc|cc}
\toprule
 & \multicolumn{4}{c|}{Dependent} & \multicolumn{2}{c}{Conventional} \\
\textbf{Method} & 1st-Sub & 2nd-Sub & 1st-Q & 2nd-Q & 1st-Q & 2nd-Q \\ \midrule
ReAct & $\underline{97.4}^{\scriptscriptstyle +1.6}_{\scriptscriptstyle -1.9}$ & $97.4^{\scriptscriptstyle +1.6}_{\scriptscriptstyle -1.9}$ & $50.0^{\scriptscriptstyle +10.0}_{\scriptscriptstyle -10.0}$ & $50.0^{\scriptscriptstyle +10.0}_{\scriptscriptstyle -10.0}$ & $50.0^{\scriptscriptstyle +10.0}_{\scriptscriptstyle -10.0}$ & $50.0^{\scriptscriptstyle +10.0}_{\scriptscriptstyle -10.0}$ \\
LangMem & $\mathbf{97.7}^{\scriptscriptstyle +1.6}_{\scriptscriptstyle -1.6}$ & $\mathbf{98.4}^{\scriptscriptstyle +1.3}_{\scriptscriptstyle -1.6}$ & $49.0^{\scriptscriptstyle +10.0}_{\scriptscriptstyle -10.0}$ & $44.0^{\scriptscriptstyle +10.0}_{\scriptscriptstyle -9.0}$ & $47.0^{\scriptscriptstyle +10.0}_{\scriptscriptstyle -10.0}$ & $49.0^{\scriptscriptstyle +10.0}_{\scriptscriptstyle -10.0}$ \\
Mem0 & $\mathbf{97.7}^{\scriptscriptstyle +1.6}_{\scriptscriptstyle -1.9}$ & $96.8^{\scriptscriptstyle +1.9}_{\scriptscriptstyle -1.9}$ & $49.0^{\scriptscriptstyle +10.0}_{\scriptscriptstyle -10.0}$ & $47.0^{\scriptscriptstyle +10.0}_{\scriptscriptstyle -10.0}$ & $53.0^{\scriptscriptstyle +10.0}_{\scriptscriptstyle -10.0}$ & $42.0^{\scriptscriptstyle +10.0}_{\scriptscriptstyle -9.0}$ \\
AWM & $96.8^{\scriptscriptstyle +1.9}_{\scriptscriptstyle -2.3}$ & $97.4^{\scriptscriptstyle +1.6}_{\scriptscriptstyle -1.9}$ & $69.0^{\scriptscriptstyle +9.0}_{\scriptscriptstyle -9.0}$ & $76.0^{\scriptscriptstyle +8.0}_{\scriptscriptstyle -9.0}$ & $\underline{54.0}^{\scriptscriptstyle +10.0}_{\scriptscriptstyle -10.0}$ & $53.0^{\scriptscriptstyle +10.0}_{\scriptscriptstyle -10.0}$ \\
DC-RS & $96.4^{\scriptscriptstyle +1.9}_{\scriptscriptstyle -2.3}$ & $\underline{97.7}^{\scriptscriptstyle +1.6}_{\scriptscriptstyle -1.6}$ & $64.0^{\scriptscriptstyle +9.0}_{\scriptscriptstyle -9.0}$ & $67.0^{\scriptscriptstyle +9.0}_{\scriptscriptstyle -10.0}$ & $\mathbf{55.0}^{\scriptscriptstyle +10.0}_{\scriptscriptstyle -10.0}$ & $\underline{57.0}^{\scriptscriptstyle +10.0}_{\scriptscriptstyle -10.0}$ \\
ExpRAG & $\underline{97.4}^{\scriptscriptstyle +1.6}_{\scriptscriptstyle -1.9}$ & $\mathbf{98.4}^{\scriptscriptstyle +1.3}_{\scriptscriptstyle -1.6}$ & $\underline{82.0}^{\scriptscriptstyle +7.0}_{\scriptscriptstyle -8.0}$ & $\underline{84.0}^{\scriptscriptstyle +7.0}_{\scriptscriptstyle -7.0}$ & $51.0^{\scriptscriptstyle +10.0}_{\scriptscriptstyle -10.0}$ & $56.0^{\scriptscriptstyle +10.0}_{\scriptscriptstyle -10.0}$ \\
ReMem & $97.1^{\scriptscriptstyle +1.6}_{\scriptscriptstyle -1.9}$ & $96.8^{\scriptscriptstyle +1.9}_{\scriptscriptstyle -2.3}$ & $76.0^{\scriptscriptstyle +8.0}_{\scriptscriptstyle -9.0}$ & $66.0^{\scriptscriptstyle +9.0}_{\scriptscriptstyle -9.0}$ &  $51.0^{\scriptscriptstyle +10.0}_{\scriptscriptstyle -10.0}$ & $55.0^{\scriptscriptstyle +10.0}_{\scriptscriptstyle -10.0}$ \\ 
\ourmethod & $96.8^{\scriptscriptstyle +1.9}_{\scriptscriptstyle -1.9}$ & $97.4^{\scriptscriptstyle +1.6}_{\scriptscriptstyle -1.9}$ & $\mathbf{90.0}^{\scriptscriptstyle +5.0}_{\scriptscriptstyle -6.0}$ & $\mathbf{89.0}^{\scriptscriptstyle +6.0}_{\scriptscriptstyle -6.0}$ & $51.0^{\scriptscriptstyle +10.0}_{\scriptscriptstyle -10.0}$ & $\mathbf{62.0}^{\scriptscriptstyle +9.0}_{\scriptscriptstyle -10.0}$ \\
\bottomrule
\end{tabular}
\end{table}

To further quantify whether different stream designs yield clearer separation among methods, we additionally perform pairwise comparisons on CodeEval-Pro. For each stream setting, we compute the pairwise accuracy difference between evaluated methods under bootstrap resampling and report two descriptive statistics: the proportion of pairwise comparisons whose $95\%$ CI excludes zero, and the mean absolute pairwise delta in percentage points. 
As shown in Table~\ref{tab:codeeval_pairwise_significance}, dependent streams yield substantially stronger pairwise separation than conventional streams. 
The same trend appears in the mean absolute delta.
The independent setting remains comparatively weakly separated, consistent with its role as a robustness check rather than a setting with explicit transfer opportunities.
These results provide complementary evidence that dependent streams make differences among memory methods more statistically visible than conventional streams.
For CodeEval-Pro, CIs are computed using bootstrapping across all three-run results. 
For BrowseComp+, CIs are computed from task-level bootstrap results across $100$ tasks.

\begin{table}[t]
\centering
\small
\caption{Statistical significance of pairwise method comparisons on CodeEval-Pro.  
The first row denotes the proportion of pairwise comparisons whose 95\% CI excludes zero.
Higher values indicate that the stream yields clearer pairwise separation among evaluated methods.}
\label{tab:codeeval_pairwise_significance}
\begin{tabular}{lcccccc}
\toprule
& \multicolumn{3}{c}{Dependent} & \multicolumn{3}{c}{Conventional} \\
\cmidrule(lr){2-4} \cmidrule(lr){5-7}
\textbf{Statistics} & 1st Pass & 2nd Pass & Indep. & 1st Pass & 2nd Pass & Indep. \\
\midrule
Pairwise CI excludes zero (\%) & $34.1$ & $31.9$ & $28.6$ & $7.1$ & $6.0$ & $17.8$ \\
Mean absolute delta (pp)       & $11.2$ & $10.3$ & \ \ $4.0$ & $4.8$ & $3.7$ & \ \ $4.2$  \\
\bottomrule
\end{tabular}
\end{table}

\subsection{Detailed Performance}
\label{subsec:agentboard-block}

\begin{table}[tbp]
\centering
\small
\setlength{\tabcolsep}{4pt}
\caption{
Accuracy (\%) on CodeEval-Pro complex tasks (BigCodeBench-Lite-Pro and HumanEval-Pro). 
BigCodeBench-Lite-Pro is used for conventional and dependent task streams. 
HumanEval-Pro is used as independent tasks after the dependent or conventional stream. 
In the self-invoking setting, each complex task is provided with its corresponding subtask as reference context, so no retrieval is required.
PG denotes the difference from the memoryless agent on the corresponding stream, and SG denotes the difference between the 2nd pass and the 1st pass. 
Each value is the average over three runs. Bold and underline mark the highest and second-highest values for each metric, respectively.
\textit{Std. Dev.} denotes the population standard deviation across available non-self-invoking methods for each column.
}
\label{tab:bigcodebench}
\setlength{\tabcolsep}{3pt}
\begin{tabular}{l|lll|lll}
\toprule
& \multicolumn{3}{c|}{Dependent Stream} & \multicolumn{3}{c}{Conventional Stream} \\
\textbf{Method} 
& 1st (PG) & 2nd (SG) & Indep.
& 1st (PG) & 2nd (SG) & Indep. \\
\midrule
Self-invoking
& $60.4$
& $60.4$
& --
& --
& --
& $79.2$ \\

\ \ w/ subtask solution
& $59.0$
& $59.0$
& --
& --
& --
& $73.3$ \\
\midrule

ReAct
& $44.8$
& $44.8$
& $\mathbf{72.5}$
& $\mathbf{44.8}$
& $\mathbf{44.8}$
& $\mathbf{72.5}$ \\

LangMem
& $44.4$\ann{$-0.4$}
& $45.8$\ann{$\underline{+1.4}$}
& $70.8$\ann{$\underline{-1.7}$}
& $39.6$\ann{$-5.2$}
& $42.4$\ann{$\underline{+2.8}$}
& $66.7$\ann{$-5.8$} \\

Mem0
& $43.8$\ann{$-1.0$}
& $44.4$\ann{$+0.6$}
& $70.8$\ann{$\underline{-1.7}$}
& $43.8$\ann{$\underline{-1.0}$}
& $42.4$\ann{$-1.4$}
& $69.2$\ann{$-3.3$} \\

AWM
& $39.6$\ann{$-5.2$}
& $38.2$\ann{$-1.4$}
& $70.0$\ann{$-2.5$}
& $\underline{44.5}$\ann{$\mathbf{-0.3}$}
& $41.0$\ann{$-3.5$}
& $70.0$\ann{$\underline{-2.5}$} \\

DC-RS
& $48.6$\ann{$+3.8$}
& $50.7$\ann{$\mathbf{+2.1}$}
& $60.8$\ann{$-11.7$}
& $35.4$\ann{$-9.4$}
& $39.6$\ann{$\mathbf{+4.2}$}
& $65.8$\ann{$-6.7$} \\

ExpRAG
& $\underline{62.5}$\ann{$\underline{+17.7}$}
& $\underline{62.5}$\ann{$+0.0$}
& $68.3$\ann{$-4.2$}
& $39.6$\ann{$-5.2$}
& $39.6$\ann{$+0.0$}
& $69.2$\ann{$-3.3$} \\

ReMem
& $58.3$\ann{$+13.5$}
& $56.3$\ann{$-2.0$}
& $67.5$\ann{$-5.0$}
& $41.7$\ann{$-3.1$}
& $43.8$\ann{$+2.1$}
& $70.0$\ann{$\underline{-2.5}$} \\

\ourmethod
& $\mathbf{66.7}$\ann{$\mathbf{+21.9}$}
& $\mathbf{64.6}$\ann{$-2.1$}
& $\underline{71.7}$\ann{$\mathbf{-0.8}$}
& $43.1$\ann{$-1.7$}
& $\underline{44.5}$\ann{$+1.4$}
& $\underline{70.8}$\ann{$\mathbf{-1.7}$} \\

\midrule
\textit{Std. Dev.}
& $9.4$
& $8.8$
& $3.5$
& $3.0$
& $1.9$
& $2.0$ \\
\bottomrule
\end{tabular}
\end{table}

\begin{table}[tbp]
\centering
\small
\setlength{\tabcolsep}{4.0pt}
\caption{Accuracy (\%) of $100$ tasks sampled from the original BrowseComp+ dataset, evaluated with two passes.}
\label{tab:browsecomp}
\begin{tabular}{l|cc|cc}
\toprule
 & \multicolumn{2}{c|}{Dependent Stream} & \multicolumn{2}{c}{Conventional Stream} \\
\textbf{Method} & 1st-Q (PG) & 2nd-Q (SG) & 1st-Q (PG) & 2nd-Q (SG) \\ 
\midrule
ReAct
& $50.0$
& $50.0$
& $50.0$
& $50.0$ \\

LangMem
& $49.0$\ann{$-1.0$}
& $44.0$\ann{$-5.0$}
& $47.0$\ann{$-3.0$}
& $49.0$\ann{$+2.0$} \\

Mem0
& $49.0$\ann{$-1.0$}
& $47.0$\ann{$-2.0$}
& $53.0$\ann{$+3.0$}
& $42.0$\ann{$-11.0$} \\

AWM
& $69.0$\ann{$+19.0$}
& $76.0$\ann{$\mathbf{+7.0}$}
& $\underline{54.0}$\ann{$\underline{+4.0}$}
& $53.0$\ann{$-1.0$} \\

DC-RS
& $64.0$\ann{$+14.0$}
& $67.0$\ann{$\underline{+3.0}$}
& $\mathbf{55.0}$\ann{$\mathbf{+5.0}$}
& $\underline{57.0}$\ann{$+2.0$} \\

ExpRAG
& $\underline{82.0}$\ann{$\underline{+32.0}$}
& $\underline{84.0}$\ann{$+2.0$}
& $51.0$\ann{$+1.0$}
& $56.0$\ann{$\underline{+5.0}$} \\

ReMem
& $76.0$\ann{$+26.0$}
& $66.0$\ann{$-10.0$}
& $51.0$\ann{$+1.0$}
& $55.0$\ann{$+4.0$} \\ 

\ourmethod
& $\mathbf{90.0}$\ann{$\mathbf{+40.0}$}
& $\mathbf{89.0}$\ann{$-1.0$}
& $51.0$\ann{$+1.0$}
& $\mathbf{62.0}$\ann{$\mathbf{+11.0}$} \\ 
\midrule
\textit{Std. Dev.}
& $14.9$
& $16.0$
& $2.3$
& $5.7$ \\
\bottomrule
\end{tabular}
\end{table}

We report the detailed results on CodeEval-Pro in Table \ref{tab:bigcodebench}, the accuracy on BrowseComp+ complex tasks in Table \ref{tab:browsecomp}, and the accuracy on BrowseComp+ subtasks in Table \ref{tab:browsecomp_sub}.

AgentBoard ScienceWorld contains task families in which tasks share the same overall goal but differ in environment configurations.
In addition to the original conventional stream, we evaluate a block stream that groups tasks from the same family together.
This setting tests repeated exposure to near-duplicate goals rather than compositional reuse.
As shown in Table~\ref{tab:scienceworld}, the block stream produces limited method separation, similar to the conventional stream.
This contrasts with the larger dispersion observed in dependent coding and deep-research streams.

\begin{table}[tbp]
\centering
\small
\setlength{\tabcolsep}{4.5pt}
\caption{Accuracy (\%) of $308$ subtasks synthesized on BrowseComp+, evaluated with two passes.}
\label{tab:browsecomp_sub}
\begin{tabular}{l|cc}
\toprule
 & \multicolumn{2}{c}{Dependent} \\
\textbf{Method} & 1st-Sub (PG) & 2nd-Sub (SG) \\ 
\midrule
ReAct
& $\underline{97.4}$
& $97.4$ \\

LangMem
& $\mathbf{97.7}$\ann{$\mathbf{+0.3}$}
& $\mathbf{98.4}$\ann{$+0.7$} \\

Mem0
& $\mathbf{97.7}$\ann{$\mathbf{+0.3}$}
& $96.8$\ann{$-0.9$} \\

AWM
& $96.8$\ann{$-0.6$}
& $97.4$\ann{$+0.6$} \\

DC-RS
& $96.4$\ann{$-1.0$}
& $\underline{97.7}$\ann{$\mathbf{+1.3}$} \\

ExpRAG
& $\underline{97.4}$\ann{$\underline{+0.0}$}
& $\mathbf{98.4}$\ann{$\underline{+1.0}$} \\

ReMem
& $97.1$\ann{$-0.3$}
& $96.8$\ann{$-0.3$} \\ 

\ourmethod
& $96.8$\ann{$-0.6$}
& $97.4$\ann{$+0.6$} \\ 
\midrule
\textit{Std. Dev.}
& $0.4$
& $0.6$ \\
\bottomrule
\end{tabular}
\end{table}

\begin{table}[tbp]
\centering
\small
\setlength{\tabcolsep}{5pt}
\caption{Performance (\%) on AgentBoard ScienceWorld. \textit{SR}: success rate. \textit{PR}: progress rate.}
\label{tab:scienceworld}
\begin{tabular}{lcccc|cccc}
\toprule
 & \multicolumn{4}{c|}{Conventional Stream} & \multicolumn{4}{c}{Block Stream} \\
\cmidrule(lr){2-5} \cmidrule(lr){6-9}
Method & 1-SR & 1-PR & 2-SR & 2-PR & 1-SR & 1-PR & 2-SR & 2-PR \\
\midrule
ReAct & $\underline{64.4}$ & $\underline{87.1}$ & $64.4$ & $87.1$ & $\mathbf{64.4}$ & $\underline{87.1}$ & $\underline{64.4}$ & $\underline{87.1}$ \\
ExpRAG & $58.9$ & $86.0$ & $\mathbf{67.8}$ & $\underline{87.3}$ & $\mathbf{64.4}$ & $\mathbf{87.8}$ & $\mathbf{65.6}$ & $86.9$ \\
ReMem & $59.3$ & $83.3$ & $62.9$ & $87.1$ & $\mathbf{64.4}$ & $85.7$ & $62.2$ & $85.0$ \\
\ourmethod & $\mathbf{66.7}$ & $\mathbf{88.6}$ & $\underline{66.7}$ & $\mathbf{88.1}$ & $\underline{60.0}$ & $85.2$ & $62.2$ & $\mathbf{87.2}$ \\ \midrule
\textit{Std. Dev.}
& $3.3$
& $1.9$
& $1.9$
& $0.4$
& $1.9$
& $1.0$
& $1.5$
& $0.9$ \\
\bottomrule
\end{tabular}
\end{table}

\begin{table}[t]
\centering
\small
\caption{
Hits@k retrieval performance on dependent streams. 
\textit{Comp. $\rightarrow$ Sub} denotes retrieving the corresponding subtask memory given a complex task query, while \textit{Comp. $\rightarrow$ Self} denotes retrieving the memory of the same complex task from the previous pass.
}
\begin{tabular}{lccc}
\toprule
\textbf{Dataset} 
& \textbf{Pass 1 Comp. $\rightarrow$ Sub} 
& \textbf{Pass 2 Comp. $\rightarrow$ Sub} 
& \textbf{Pass 2 Comp. $\rightarrow$ Self} \\
\midrule
BigCodeBench-Lite-Pro & $95.8$ & $91.7$ & $98.8$ \\
BrowseComp+           & $87.0$ & $77.0$ & $98.0$ \\
\bottomrule
\end{tabular}

\label{tab:retrieval_hits}
\end{table}

Since most evaluated memory methods distill memories from one or multiple tasks, and the resulting memory states may be updated over time, directly evaluating the retrieval step is less straightforward. 
Therefore, we primarily report the retrieval performance of \ourmethod on the dependent streams of our two main datasets, as shown in Table~\ref{tab:retrieval_hits}. We measure Hits@k, defined as the proportion of the top-$2$ retrieved memory entries that correspond to relevant and reusable experience. 
The results show that retrieving the memory of the corresponding previous task is not a major bottleneck, suggesting that the more critical challenge lies in how memories are represented and utilized. 
We also observe that Complex $\rightarrow$ Sub of the second pass is lower than that of the first pass on both datasets. 
This is expected because Complex $\rightarrow$ Self memories in the second pass provide more directly relevant experience and thus compete with subtask memories during retrieval. 
Importantly, a high hit rate of Complex $\rightarrow$ Self does not constitute answer leakage: memories must be constructed by the agent itself in our setting, and the previous attempt is only available as memory context, serving as a reference rather than a guaranteed correct solution.

\subsection{Performance Across Backbone Models}
\label{subsec:backbone-robustness}

We replicate experiments on CodeEval-Pro using open-weight gpt-oss-120b and proprietary Claude Sonnet 5, while keeping all other settings fixed.
Table~\ref{tab:backbone-robustness} compares these results with the Qwen3.5-35B-A3B backbone presented in the main text.
Despite different backbone models, our main conclusions remain.

\begin{table}[tbp]
\centering
\small
\setlength{\tabcolsep}{5pt}
\caption{Accuracy (\%) on CodeEval-Pro complex tasks using three backbone models. Each stream is evaluated with two passes; PG and SG are shown in parentheses. The Qwen3.5-35B-A3B results reproduce the corresponding three-run averages in Table~\ref{tab:bigcodebench}.}
\label{tab:backbone-robustness}
\begin{tabular}{lcccc}
\toprule
& \multicolumn{2}{c}{Dependent Stream} & \multicolumn{2}{c}{Conventional Stream} \\
\cmidrule(lr){2-3}\cmidrule(lr){4-5}
\textbf{Method} & 1st (PG) & 2nd (SG) & 1st (PG) & 2nd (SG) \\
\midrule
\rowcolor{gray!12}\multicolumn{5}{c}{\textit{Qwen3.5-35B-A3B}} \\
ReAct & $44.8$ & $44.8$ & $44.8$ & $44.8$ \\
ExpRAG & $62.5$\ann{$+17.7$} & $62.5$\ann{$+0.0$} & $39.6$\ann{$-5.2$} & $39.6$\ann{$+0.0$} \\
ReMem & $58.3$\ann{$+13.5$} & $56.3$\ann{$-2.0$} & $41.7$\ann{$-3.1$} & $43.8$\ann{$+2.1$} \\
\ourmethod & $66.7$\ann{$+21.9$} & $64.6$\ann{$-2.1$} & $43.1$\ann{$-1.7$} & $44.5$\ann{$+1.4$} \\
\midrule
\rowcolor{gray!12}\multicolumn{5}{c}{\textit{gpt-oss-120b}} \\
ReAct & $45.8$ & $45.8$ & $45.8$ & $45.8$ \\
ExpRAG & $58.3$\ann{$+12.5$} & $58.3$\ann{$+0.0$} & $37.5$\ann{$-8.3$} & $35.4$\ann{$-2.1$} \\
ReMem & $56.3$\ann{$+10.4$} & $52.1$\ann{$-4.2$} & $43.8$\ann{$-2.1$} & $43.8$\ann{$+0.0$} \\
\ourmethod & $52.1$\ann{$+6.3$} & $56.3$\ann{$+4.2$} & $35.4$\ann{$-10.4$} & $33.3$\ann{$-2.1$} \\
\midrule
\rowcolor{gray!12}\multicolumn{5}{c}{\textit{Claude Sonnet 5}} \\
ReAct & $50.0$ & $50.0$ & $50.0$ & $50.0$ \\
ExpRAG & $62.5$\ann{$+12.5$} & $64.6$\ann{$+2.1$} & $47.9$\ann{$-2.1$} & $47.9$\ann{$+0.0$} \\
ReMem & $68.8$\ann{$+18.8$} & $64.6$\ann{$-4.2$} & $47.9$\ann{$-2.1$} & $45.8$\ann{$-2.1$} \\
\ourmethod & $58.3$\ann{$+8.3$} & $56.3$\ann{$-2.0$} & $50.0$\ann{$+0.0$} & $50.0$\ann{$+0.0$} \\
\bottomrule
\end{tabular}
\end{table}

\subsection{Sensitivity to Retrieval Budget}
\label{subsec:retrieval-budget}

We additionally evaluate \ourmethod with $k\in\{1,2,4\}$, where $k$ is the number of retrieved chunks during inference, keeping all other settings fixed.
As shown in Table~\ref{tab:retrieval-budget}, the contrast between dependent and conventional streams remains regardless of the choices of $k$.
Increasing $k$ to $4$ does not monotonically improve performance and slightly worsens conventional-stream plasticity, suggesting that retrieving more potentially irrelevant experience can introduce interference.

\begin{table}[tbp]
\centering
\small
\setlength{\tabcolsep}{5pt}
\caption{Effect of retrieval depth $k$ on \ourmethod for CodeEval-Pro complex tasks. Accuracy (\%) is reported with PG and SG in parentheses. 1st and 2nd denote the first and second passes over the same stream.}
\label{tab:retrieval-budget}
\begin{tabular}{ccccc}
\toprule
& \multicolumn{2}{c}{Dependent Stream} & \multicolumn{2}{c}{Conventional Stream} \\
\cmidrule(lr){2-3}\cmidrule(lr){4-5}
$k$ & 1st (PG) & 2nd (SG) & 1st (PG) & 2nd (SG) \\
\midrule
$1$ & $66.7$\ann{$+21.9$} & $60.4$\ann{$-6.3$} & $45.8$\ann{$+1.0$} & $45.8$\ann{$+0.0$} \\
$2$ & $66.7$\ann{$+21.9$} & $64.6$\ann{$-2.1$} & $43.1$\ann{$-1.7$} & $44.5$\ann{$+1.4$} \\
$4$ & $64.6$\ann{$+19.8$} & $64.6$\ann{$+0.0$} & $41.7$\ann{$-3.1$} & $43.8$\ann{$+2.1$} \\
\bottomrule
\end{tabular}
\end{table}

\subsection{Sensitivity to Source--Target Distance}
\label{subsec:task-distance}

We examine whether transfer degrades as the distance between a subtask and its dependent complex task grows, i.e., whether there is a forgetting window for the evaluated memory designs.
The dependent streams already contain substantial natural variation in this source-target task distance.
On CodeEval-Pro, the $48$ subtasks and the $48$ complex tasks are shuffled independently before the two blocks are concatenated, so the number of intervening tasks between a subtask and its dependent task ranges from $7$ to $86$, with a mean of $47$.
We partition this range into four equal-width intervals and report accuracy, PG, and task retrieval Hits@2 within each interval.
PG is measured against the memoryless ReAct result on the same tasks, which partly controls for differences in task difficulty across intervals.

As shown in Table~\ref{tab:task-distance}, performance does not degrade monotonically with distance.
In the longest-distance interval, where the source and target tasks are separated by $75.7$ tasks on average, all three memory designs retain a PG of $+18.2$, and the relevant subtask is still retrieved in $90.9\%$--$100\%$ of cases.
We therefore find no evidence of an effective forgetting window at the stream lengths, at least within the range observed in our experiments.

\begin{table}[tbp]
\centering
\small
\setlength{\tabcolsep}{5pt}
\caption{Accuracy (\%), PG (in parentheses), and task retrieval Hits@2 (\%) on the CodeEval-Pro dependent stream (first pass), grouped by task distance (the number of intervening tasks between a subtask and its dependent complex task). PG is computed against ReAct on the same tasks within each interval.}
\label{tab:task-distance}
\begin{tabular}{lcccc}
\toprule
\textbf{Task Distance} & $7$--$26$ & $27$--$46$ & $47$--$66$ & $67$--$86$ \\
\midrule
\# Task pairs & $9$ & $16$ & $12$ & $11$ \\
Mean distance & $19.2$ & $35.9$ & $56.3$ & $75.7$ \\
\midrule
\rowcolor{gray!12}\multicolumn{5}{c}{\textit{Accuracy (PG)}} \\
ReAct & $33.3$ & $68.8$ & $33.3$ & $36.4$ \\
ExpRAG & $44.4$\ann{$+11.1$} & $81.2$\ann{$+12.5$} & $58.3$\ann{$+25.0$} & $54.5$\ann{$+18.2$} \\
ReMem & $44.4$\ann{$+11.1$} & $75.0$\ann{$+6.2$} & $50.0$\ann{$+16.7$} & $54.5$\ann{$+18.2$} \\
\ourmethod & $44.4$\ann{$+11.1$} & $81.2$\ann{$+12.5$} & $58.3$\ann{$+25.0$} & $54.5$\ann{$+18.2$} \\
\midrule
\rowcolor{gray!12}\multicolumn{5}{c}{\textit{Hits@2}} \\
ExpRAG & $100.0$ & $100.0$ & $91.7$ & $100.0$ \\
ReMem & $88.9$ & $93.8$ & $100.0$ & $90.9$ \\
\ourmethod & $100.0$ & $93.8$ & $91.7$ & $100.0$ \\
\bottomrule
\end{tabular}
\end{table}

\subsection{Sensitivity to Stream Position}
\label{subsec:stream-position}

We examine how PG evolves as a stream proceeds, i.e., whether memory methods improve at later positions as more experience is accumulated.
To this end, we partition each stream into consecutive position bins without resetting memory, and report PG within each bin, together with its average across methods.

Table~\ref{tab:stream-position} shows that the trend depends on the stream type.
On conventional streams, PG does not increase as more tasks are processed, and often declines.
Averaged across methods, PG drops from $+8.5$ in the first bin to $+0.5$ in the last on MMLU-Pro, from $-1.1$ to $-3.3$ on ScienceWorld, and ends at $-8.3$ on CodeEval-Pro.
Accumulating weakly related experience therefore does not translate into improved plasticity.
In contrast, on the CodeEval-Pro dependent stream, PG is positive in every bin and increases toward later positions, with the method average rising from $+8.3$ in the first bin to $+22.2$ in the last.
Since both CodeEval-Pro streams evaluate the same complex tasks in the same order and bins, this contrast is not due to task difficulty.
Together with the distance analysis in Appendix~\ref{subsec:task-distance}, these results suggest that whether memory benefits from a longer stream depends on the presence of reusable task relations.

\begin{table}[tbp]
\centering
\small
\setlength{\tabcolsep}{5pt}
\caption{PG (\%) by task position, with memory accumulated continuously across each stream. Bins are consecutive and of equal size: CodeEval-Pro $4\times12$ complex tasks, MMLU-Pro $4\times25$ tasks per domain, and ScienceWorld $3\times30$ tasks. The two CodeEval-Pro streams share the same complex tasks in the same order and bins; in the dependent stream, all $48$ subtasks precede the complex tasks. \textit{Average} denotes the mean over the three methods. PG is computed against ReAct on the same tasks within each bin. Results are from a single run; Table~\ref{tab:bigcodebench} reports three-run averages.}
\label{tab:stream-position}
\begin{tabular}{lcccc}
\toprule
\textbf{Method} & Bin 1 & Bin 2 & Bin 3 & Bin 4 \\
\midrule
\rowcolor{gray!12}\multicolumn{5}{c}{\textit{CodeEval-Pro, conventional stream}} \\
ExpRAG & $-8.3$ & $-8.3$ & $+8.3$ & $-16.7$ \\
ReMem & $+0.0$ & $-8.3$ & $+0.0$ & $-8.3$ \\
\ourmethod & $+0.0$ & $-16.7$ & $+0.0$ & $+0.0$ \\
\cmidrule(lr){1-5}
\textit{Average} & $-2.8$ & $-11.1$ & $+2.8$ & $-8.3$ \\
\midrule
\rowcolor{gray!12}\multicolumn{5}{c}{\textit{CodeEval-Pro, dependent stream}} \\
ExpRAG & $+8.3$ & $+8.3$ & $+25.0$ & $+25.0$ \\
ReMem & $+8.3$ & $+0.0$ & $+25.0$ & $+16.7$ \\
\ourmethod & $+8.3$ & $+25.0$ & $+8.3$ & $+25.0$ \\
\cmidrule(lr){1-5}
\textit{Average} & $+8.3$ & $+11.1$ & $+19.4$ & $+22.2$ \\
\midrule
\rowcolor{gray!12}\multicolumn{5}{c}{\textit{MMLU-Pro, conventional stream}} \\
ExpRAG & $+10.7$ & $-2.7$ & $+2.7$ & $+0.0$ \\
ReMem & $+6.7$ & $-4.0$ & $+0.0$ & $-1.3$ \\
\ourmethod & $+8.0$ & $+4.0$ & $+1.3$ & $+2.7$ \\
\cmidrule(lr){1-5}
\textit{Average} & $+8.5$ & $-0.9$ & $+1.3$ & $+0.5$ \\
\midrule
\rowcolor{gray!12}\multicolumn{5}{c}{\textit{ScienceWorld, conventional stream}} \\
ExpRAG & $-3.3$ & $-6.7$ & $-6.7$ & -- \\
ReMem & $-3.3$ & $-6.7$ & $-3.3$ & -- \\
\ourmethod & $+3.3$ & $+3.3$ & $+0.0$ & -- \\
\cmidrule(lr){1-5}
\textit{Average} & $-1.1$ & $-3.4$ & $-3.3$ & -- \\
\bottomrule
\end{tabular}
\end{table}

\subsection{Details in Benchmark Construction}
\label{subsec:details-in-benchmark-construction}

We provide details on the construction of the AgentBoard BabyAI dataset.
We constructed two task streams from this dataset. 
It originally had $112$ tasks, organized into $28$ BabyAI families, with four tasks per family. 
Tasks in each family share the same overall goal but differ in their environments.
For the dependent stream, we selected seven high-confidence source-to-target task-family relations.
Each relation maps a simpler source family to a target family that requires an additional or composed skill, such as single-object navigation to sequential navigation, navigation to pickup, single-door opening to ordered two-door opening, or pickup to unlock-then-pickup.
For each selected pair, we kept all four source tasks and all four target tasks. 
The $28$ source tasks and $28$ target tasks were shuffled independently and randomly, then concatenated as a source half followed by a target half. 
This yields a deterministic $56$-task bipartite stream designed to test whether earlier experience with simpler task families can transfer to later, compositionally related target families. 
Given that a special property of AgentBoard BabyAI is that its task families exhibit clear relations, which have already been exploited in constructing the dependent stream, we aim to exclude the influence of such relations when constructing the conventional stream.
We selected ten easy-only BabyAI families covering diverse atomic skills, including navigation, target disambiguation, door opening, pickup, obstacle-aware pickup, location-grounded pickup, unlocking, and unlock-then-pickup. 
Each selected family was required to contain exactly $4$ tasks. 
Families were ordered according to their first appearance in the original split, and within-family order was preserved, which yields a $40$-task stream.

Table~\ref{tab:babyai-heldout} reports results on $36$ held-out AgentBoard BabyAI tasks that occur in neither stream, using memory constructed separately on each stream and frozen before evaluation. \ourmethod improves both SR and PR over memoryless ReAct after either stream.

\begin{table}[tbp]
\centering
\small
\setlength{\tabcolsep}{5pt}
\caption{Performance (\%) on $36$ held-out AgentBoard BabyAI tasks using frozen memory constructed from the dependent or conventional stream. \textit{SR}: success rate. \textit{PR}: progress rate.}
\label{tab:babyai-heldout}
\begin{tabular}{lcccc}
\toprule
& \multicolumn{2}{c}{Dependent Stream} & \multicolumn{2}{c}{Conventional Stream} \\
\cmidrule(lr){2-3}\cmidrule(lr){4-5}
\textbf{Method} & SR & PR & SR & PR \\
\midrule
ReAct & $\underline{36.1}$ & $\underline{58.8}$ & $36.1$ & $58.8$ \\
ExpRAG & $30.6$ & $53.8$ & $\mathbf{47.2}$ & $60.7$ \\
ReMem & $\underline{36.1}$ & $53.2$ & $41.7$ & $\underline{61.9}$ \\
\ourmethod & $\mathbf{41.7}$ & $\mathbf{64.8}$ & $\underline{44.4}$ & $\mathbf{63.1}$ \\
\bottomrule
\end{tabular}
\end{table}

\subsection{Case Study}
\label{appendix-subsec:case-study}

We show two coding cases in Table \ref{tab:case_study}, as discussed in \S\ref{subsec:case-study}.

\begin{table}[t]
\centering
\small
\setlength{\tabcolsep}{0pt}
\caption{Case study on coding tasks. Retrieved memory context helps when it captures the reusable subtask structure, but hurts when a topically similar task has incompatible semantics.}
\begin{tabularx}{\linewidth}{@{}Y@{}}
\toprule

\textbf{Positive transfer: correct compositional relation}
(BigCodeBench/516; \ourmethod{} pass, ReAct fail) \\
\midrule

\textbf{Target decisive condition.}
The target function receives multiple pairs of numeric lists and must return three lists: Euclidean distances, DataFrames, and Axes objects. The decisive test checks that each DataFrame has shape $(n,2)$ with columns \texttt{A} and \texttt{B}, e.g., \texttt{pd.DataFrame(\{'A': [1,2,3], 'B': [4,5,6]\})}. \\[2pt]

\textbf{Baseline behavior.}
The baseline misinterprets the DataFrame format and creates a one-column DataFrame with repeated row indices \texttt{A} and \texttt{B}, yielding shape $(2n,1)$ instead of $(n,2)$. The test fails with a shape mismatch: predicted $(6,1)$ vs. expected $(3,2)$. \\[2pt]

\textbf{Retrieved memory.}
The retrieved task is the single-pair version of the same computation: given two lists \texttt{a} and \texttt{b}, compute \texttt{distance.euclidean(a,b)}, construct \texttt{pd.DataFrame(\{'A': a, 'B': b\})}, plot the two series, and return \texttt{(distance, df, ax)}. \\[2pt]

\textbf{Inter-task relation.}
The target is a list-level composition of the retrieved task: the same single-pair operation should be applied independently to each pair, and the corresponding outputs should be collected. \\[2pt]

\textbf{\ourmethod{} behavior.}
The memory agent reuses the retrieved DataFrame construction, \texttt{pd.DataFrame(\{'A': a, 'B': b\})}, inside a loop over pairs. This preserves the required $(n,2)$ structure and produces the expected output lists. \\

\midrule

\textbf{Negative transfer: superficial task similarity}
(BigCodeBench/877; \ourmethod{} fail, ReAct pass) \\
\midrule

\textbf{Target decisive condition.}
The target function must combine numeric DataFrames, return a covariance matrix and a Seaborn pair plot, and reject invalid inputs. The decisive test specifically passes a DataFrame with one non-numeric column, \texttt{\{'A': ['a','b'], 'B': [3,4]\}}, and expects a \texttt{TypeError}. \\[2pt]

\textbf{Baseline behavior.}
The baseline checks every column with \texttt{is\_numeric\_dtype}. It detects the non-numeric column \texttt{A} and raises \texttt{TypeError}, matching the target requirement. \\[2pt]

\textbf{Retrieved memory.}
The retrieved task also processes a list of DataFrames, but its goal is different: fill missing numeric values, standardize numeric columns, compute correlation matrices, and draw heatmaps. \\[2pt]

\textbf{Inter-task relation.}
The memory is topically similar but semantically mismatched. In the retrieved task, non-numeric columns can be ignored by selecting numeric columns. In the target task, any non-numeric column makes the input invalid. \\[2pt]

\textbf{\ourmethod{} behavior.}
The memory agent follows the retrieved pattern and uses \texttt{select\_dtypes(include=[np.number])}. For the mixed DataFrame, it keeps column \texttt{B}, drops non-numeric column \texttt{A}, and continues to compute covariance and pair plots instead of raising \texttt{TypeError}. \\

\bottomrule
\end{tabularx}
\label{tab:case_study}
\end{table}

\section{License of Existing Assets}
\label{sec:license-of-existing-assets}

We list the utilized assets and their licenses.
The released dataset artifacts of CodeEval-Pro are labeled MIT. 
The BrowseComp-Plus dataset release is labeled MIT.
The MMLU-Pro dataset is labeled MIT, while its official GitHub repository is Apache-2.0. 
ScienceWorld is Apache-2.0. 
These releases are usable for our academic research.

\section{New Assets}
\label{sec:new-assets}

The only newly generated tasks are the synthesized BrowseComp+ subtasks. We also release derived benchmark artifacts, including stream orders and evaluation splits constructed from existing datasets.

\subsection{Synthesis of BrowseComp+ Subtasks}
\label{subsec:synthesis-of-bc-subtask}

BrowseComp+ is a challenging deep research benchmark with complex, artificially constructed multi-hop information-seeking tasks. 
Because independently sampled complex tasks may share little reusable structure, directly streaming them does not provide a continuous deep research workflow with reusable knowledge.
We therefore construct a dependent stream whose purpose is to create known transfer opportunities for evaluation.
Concretely, we use the original BrowseComp+ annotations only during offline benchmark construction to identify localized, evidence-supported intermediate questions for each parent task. 
We use GPT-5.2 as an LLM generator. 
It is given the parent task together with its final answer and supporting gold/evidence documents, and is asked to propose self-contained intermediate subtasks only when it believes a faithful decomposition is possible. 
This privileged information is never exposed to the evaluated agents or memory methods at test time. 
Each generated subtask is paired with a concise reference answer and supporting document identifiers solely for dataset verification and automatic evaluation. 
We retain a candidate only if its answer is evidence-supported, deterministic, and logically relevant to the parent task. 
Although the generation prompt encourages decompositions of $2$--$5$ steps, this cardinality is not hard-enforced. 
The final stream keeps only the verified subset that survives filtering.

We enforce quality control of generated subtasks in two stages. First, we apply deterministic filtering to remove malformed candidates. A candidate subtask is retained only if both the subtask and its answer are non-empty, all cited document identifiers come from the original gold/evidence pools, both gold documents and evidence documents are non-empty, and the cited gold documents are a subset of the cited evidence documents. 
By default, we further require the proposed answer to appear verbatim in the supporting documents, which suppresses unsupported paraphrases and hallucinated answers. Second, each surviving candidate is re-verified by the model with a separate verification prompt conditioned on the original task and the cited supporting documents. The verifier checks five properties: explicit entity specification, logical relevance to the original task, evidence support for the required reasoning step, determinism of the answer, and answer completeness. 
Only candidates that pass all verifications are retained as final subtasks. 
Tasks for which the model declines decomposition, returns an invalid response, or yields no valid candidate after filtering are skipped.

Our focus is on evidence sharing within multi-hop logical chains, rather than on using subtasks to directly prompt the original task.
Entities, timestamps, and other information mentioned in the original task are not required to appear in the subtasks.

\begin{table}[tb]
\centering
\small
\caption{The statistics of subtasks and tasks on BigCodeBench-Lite-Pro. Each cell contains the mean and standard deviation.}
\label{tab:code_stats}
\begin{tabular}{lrr}
\toprule
\textbf{Statistics} & \textbf{Subtask} & \textbf{Task} \\
\midrule
Query tokens  & $304.8 \pm 118.1$ & $121.3 \pm 34.4$ \\
Solution tokens & $100.7 \pm 46.2$  & $87.2 \pm 53.6$ \\
Solution Lines  & $11.9 \pm 5.7$    & $10.7 \pm 5.2$ \\
Function Calls in Solution  & $7.4 \pm 3.3$     & $5.9 \pm 4.0$ \\
\bottomrule
\end{tabular}
\end{table}

\begin{table}[tb]
\centering
\small
\caption{The statistics of subtasks and tasks on BrowseComp+. Each cell contains the mean and standard deviation.}
\label{tab:browse_stats}
\begin{tabular}{lrr}
\toprule
\textbf{Statistics} & \textbf{Subtask} & \textbf{Task} \\
\midrule
Query Tokens & $29.7 \pm 9.9$ & $72.5 \pm 12.0$ \\
Number of evidence documents & $1.8 \pm 0.8$ & $5.0 \pm 2.1$ \\
Tokens in evidence documents & $6.3\mathrm{k} \pm 24.0\mathrm{k}$ & $22.7\mathrm{k} \pm 52.0\mathrm{k}$ \\
Query-token recall in evidence documents & $83.3\% \pm 12.4\%$ & $73.5\% \pm 14.8\%$ \\
Answer-token recall in evidence documents & $99.9\% \pm 1.9\%$ & $99.2\% \pm 6.0\%$ \\
\bottomrule
\end{tabular}
\end{table}

\subsection{The Statistics of Dependent Streams}
\label{subsec:statistics-of-compositional-streams}

Both Table~\ref{tab:code_stats} and Table~\ref{tab:browse_stats} compare the statistics of subtasks and complex tasks, revealing a shared compositional pattern in which task-level instances aggregate more information or complexity than their constituent subtasks.
In BigCodeBench-Lite-Pro, subtasks are more verbose than full tasks in terms of query length, and they also involve slightly longer solutions and more function calls. 
This suggests that the subtasks have a smaller problem scope but include more implementation details, whereas the complex tasks are often wrappers around these smaller problems.
In contrast, BrowseComp+ exhibits the opposite trend for query length, where complex tasks are longer and require more evidence documents and many more supporting tokens than subtasks, indicating that composition primarily increases retrieval breadth and evidence integration demands.
Across both coding and deep-research tasks, the transition from subtasks to tasks is associated with a change in information organization rather than a simple uniform scaling of all statistics.

We further examine whether the synthesized BrowseComp+ subtasks collapse into paraphrases of their parent task. 
The token-set Jaccard similarity between subtasks and their parent task is low, with a mean of $0.10$, and no pair exceeds $0.30$. 
Only $5.8\%$ of individual subtasks cover the full parent evidence document set, and only $1.6\%$ simultaneously have the same normalized answer and the same evidence set as the parent task. 
These diagnostics suggest that BrowseComp+ subtasks are semantically related localized evidence steps rather than full-task paraphrases or near-duplicate restatements of the complex tasks.



\end{document}

%% file: math_commands.tex
\usepackage{amsmath,amsfonts,bm}

\def\eqref#1{equation~\ref{#1}}

\def\1{\bm{1}}

\DeclareMathAlphabet{\mathsfit}{\encodingdefault}{\sfdefault}{m}{sl}
\SetMathAlphabet{\mathsfit}{bold}{\encodingdefault}{\sfdefault}{bx}{n}

